\documentclass[12pt]{article}

\usepackage[margin=1in]{geometry}
\usepackage{times}
\usepackage{latexsym}
\usepackage{amsmath}
\usepackage{amssymb}

\usepackage{xcolor}
\usepackage{tikz}
\usetikzlibrary{arrows.meta,positioning,calc,fit,backgrounds,decorations.pathreplacing}

\definecolor{ink}{HTML}{1F2428}
\definecolor{mid}{HTML}{6B7075}
\definecolor{light}{HTML}{E9ECEF}
\definecolor{paper}{HTML}{FAFAF7}
\definecolor{accent}{HTML}{2F6F73}
\definecolor{accentlight}{HTML}{DCECEA}
\definecolor{warm}{HTML}{F1EEE7}

\usepackage[T1]{fontenc}
\usepackage[utf8]{inputenc}
\usepackage{microtype}
\usepackage{textcomp}
\usepackage{textgreek}
\usepackage{inconsolata}

\usepackage{setspace}

\usepackage{multirow}
\usepackage{array}
\usepackage{graphicx}
\graphicspath{{./}}
\usepackage{subcaption}
\usepackage{booktabs}
\usepackage{natbib}
\usepackage{xcolor}
\usepackage[hyphens,spaces,obeyspaces]{url}
\usepackage{float}
\usepackage{placeins}
\usepackage{caption}
\usepackage{tabularx}
\usepackage{authblk}

\usepackage{listings}
\usepackage{titlesec}

\titleformat{\paragraph}
  {\normalfont\normalsize\bfseries}
  {}{0pt}{}

\titlespacing*{\paragraph}
  {0pt}{1.0ex}{0.8ex}

\newcommand{\figurenote}[2][0.95\linewidth]{%
  \vspace{0.35em}\par\noindent
  \parbox{#1}{\footnotesize\emph{Note.} #2}%
}

\usepackage[colorlinks=true,linkcolor=blue!60!black,citecolor=blue!60!black,urlcolor=blue!60!black]{hyperref}

\title{ORQA: An Occupation-Realistic Question and Answer Framework for LLM Professional Knowledge}

\author[1]{Shreyas Krishnan\footnote{Corresponding author: \href{mailto:shreyas.krishnan@berkeley.edu}{shreyas.krishnan@berkeley.edu}. The reproducibility code is available at \url{https://github.com/sk160902/ORQA_pipeline}. All errors are our own.}}
\author[1]{Serina Chang}
\author[1,2]{Abhishek Nagaraj}

\affil[1]{University of California, Berkeley}
\affil[2]{National Bureau of Economic Research (NBER)}

\date{September, 2026}

\begin{document}
\renewcommand*{\thefootnote}{\fnsymbol{footnote}}
\maketitle
\renewcommand*{\thefootnote}{\arabic{footnote}}
\setcounter{footnote}{0}
\begin{abstract}

We present ORQA, a method for testing occupation-level knowledge in large language models. Prior methods either map abstract LLM skills to occupations via task definitions or utilize expert knowledge which is difficult to obtain at scale and expensive. ORQA complements both of these methods by connecting O*NET occupations to trusted occupation-specific websites (such as regulatory agencies, licensing bodies, professional organizations, and government publications) and converting these into source-traceable question-answer pairs. A combination of an automated pipeline and human review produces a set of high quality questions about occupations. The question set created via our method covers 116 occupations from all 21 major groups in the SOC, with 480 questions sourced from 187 different websites. Each question is designed to probe a real-world skill question that is relevant to the occupation in question. We test 15 state-of-the-art frontier and open-weight models via this method. Claude Opus 4.6, GPT-5.4 and Claude Sonnet 4.6 all perform the best at approximately 58-62\% while smaller open-weight models achieve approximately 33-41\% performance. Performance varies significantly across occupations. Healthcare-related occupations achieve the highest performance (78\%) while Office and Administrative Support achieve approximately 40\%. Performance on individual occupations (e.g. Sheet Metal Workers and Fish and Game Wardens) is essentially zero. We also find that open-ended questions and weighting by wage bill do not significantly affect the ranking of models on this benchmark. We believe that leveraging existing trusted occupation-specific information to test LLM knowledge in professional domains may be a scalable and useful method for evaluating occupation-level AI performance in the future. Results and data are available at \href{https://orqabench.org}{orqabench.org}.

\end{abstract}

\newpage

\setstretch{1.5}

\section{Introduction}\label{sec:intro}

Benchmarks have been central to measuring and directing progress in artificial intelligence. For language models, benchmarks such as MMLU, BIG-bench, HELM, GPQA, ARC, and SWE-bench have made it possible to compare systems on reasoning, knowledge, coding, and problem solving in a reproducible way \citep{hendrycks2021mmlu,srivastava2023bigbench,liang2023helm,rein2024gpqa,chollet2019measure,jimenez2024swebench}. As language models become general-purpose work tools, however, the evaluation question shifts. We no longer only want to know whether a model has an abstract capability. We want to understand its practical impacts across the economy. We want to know whether it can help a worker in a particular occupation answer the kinds of questions that arise in practice, and whether one model is more useful than another for that occupation.

Existing work does not yet provide this kind of occupation-level model benchmark across the entire US economy across models in a scalable way. Existing economic work measures exposure, adoption, or productivity effects of AI in the labor market \citep{brynjolfsson2018machines,eloundou2023gpts,handa2025economictasks,chatterji2025chatgptusage,brynjolfsson2025genaiwork,dellacqua2026jagged}. Some papers on AI in the labor market are valuable for examining the impact of AI on the labor market. However, these are typically not performance comparison benchmarks for AI systems. Instead, these papers typically estimate the overall impact of a particular AI system on the labor market rather than comparing the usefulness of a system for specific jobs to another system for the same jobs.

\begin{figure}[!t]
  \centering
  \caption{Construction pipeline for ORQA from occupation sampling to verified benchmark questions.}
  \includegraphics[width=\linewidth]{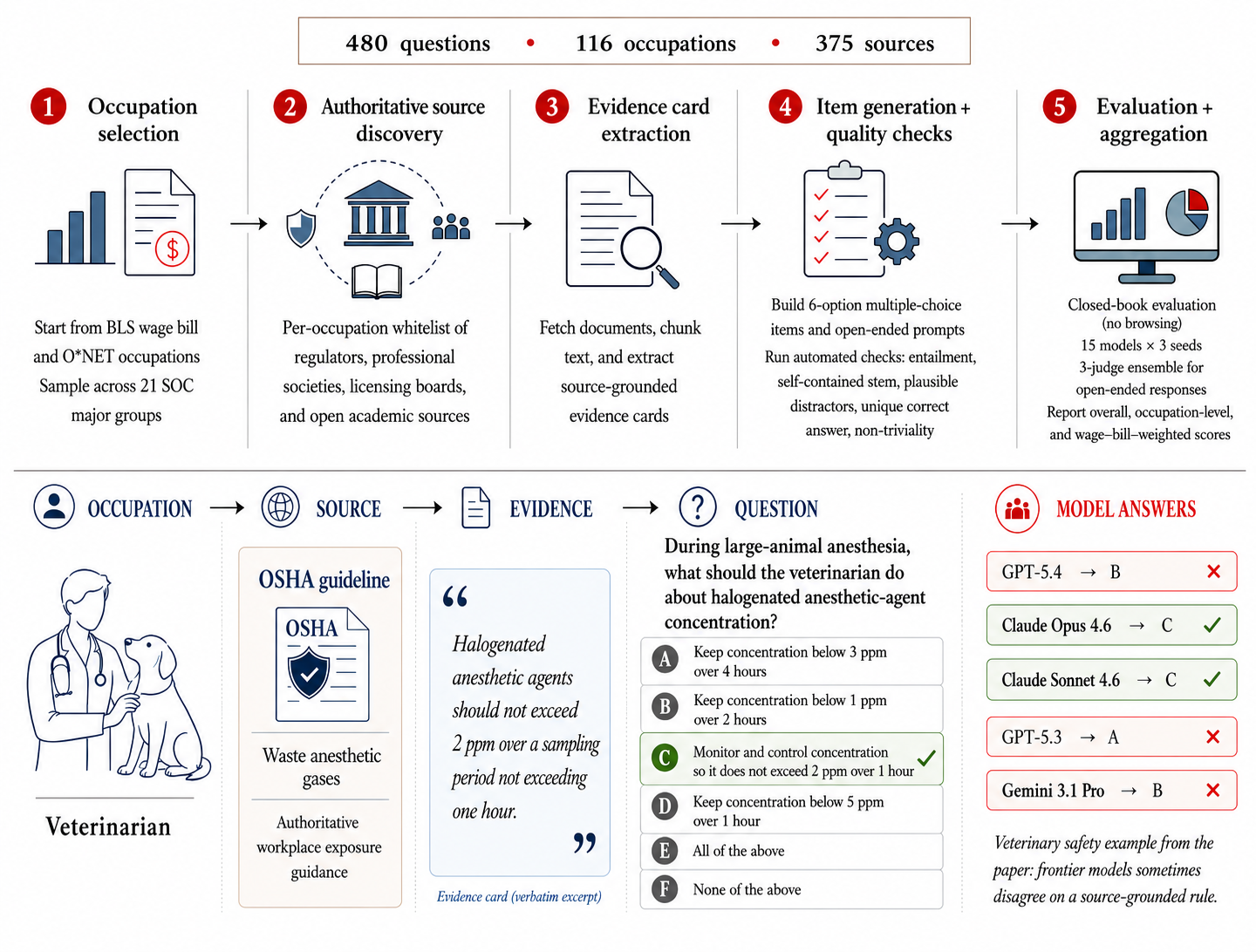}
  \figurenote{The figure illustrates the pipeline from occupation sampling and source identification to evidence-card extraction, question construction, quality assurance, and finally to the creation of a balanced benchmark composed of verified questions.}
  \label{fig:method}
\end{figure}

Several recent benchmarks have started to aim at more real tasks. These include web agent tasks \citep{zhou2023webarena}, enterprise workflows \citep{drouin2024workarena}, consequential workplace tasks \citep{xu2025agentcompany}, and software engineering tasks \citep{jimenez2024swebench}. These are promising developments. However, these benchmarks are still quite limited with respect to the labor market. Many benchmarks currently examine computer mediated digital jobs. In addition, recent attempts to align agent benchmarks to jobs have found a strong bias toward mathematical and computer- related jobs \citep{wang2025humanworkflows,wang2026agentwork}. GDPval represents an important recent effort toward examining more economically relevant tasks. However, GDPval uses expert written prompts and expert or model evaluations which limit the ability to extend to hundreds of jobs \citep{openai2025gdpval}. There have also been attempts to bridge this gap using expert evaluations of AI performance on job-task levels \citep{shao2025future}. Thus the gap appears to be one of benchmarking. We desire occupation-level assessments that are (a) scalable, (b) broadly applicable throughout the economy, and (c) capable of assessing knowledge at the level of occupations. We also desire some of the positive aspects of the current system including source verifiability and model comparisons while moving beyond purely task completion to encompass professional knowledge.

We introduce Occupation Related Question and Answer (ORQA) to fulfill this need but also to explore the feasibility of automatically creating an occupation-level benchmark that is grounded in the economy and to compare the advantages and disadvantages of this approach versus expert created benchmarks. The creation of ORQA stems from the recognition that there exists authoritative text for many occupations that contains professional knowledge relevant to professionals in that occupation. Practical knowledge that professionals are expected to utilize is found in occupation specific documents, government agencies, regulators, licensing bodies, academic sources, standards organizations, and formal and informal professional associations. This also reflects sociological understanding regarding the creation of professional norms through occupations and their codification in formal and informal sources \citep{abbott2014system}. We believe such documents can be used to provide scalable occupation-level comparisons, provided they can be connected to O*NET occupations and transformed into source-traceable questions. We are also cognizant of the scope of ORQA. Specifically, ORQA addresses the source-traceable aspect of occupational knowledge (reporting responsibilities, professional recommendations, thresholds, procedures and guidelines) rather than overall occupational capability. Furthermore, ORQA is designed to operate on a number of occupations where authoritative information exists, rather than across the entire labor market. Instead, we view ORQA as a system that demonstrates the feasibility of automated, source- traceable, occupation-level comparisons and augments both task and expert-based methods.

We propose an agentic system for the large-scale creation of occupation-related QA pairs. We begin with occupation information from the BLS and O*NET \citep{bls2024oews,onetdatabase} and create occupation-specific lists of authoritative sources. Using these lists, we retrieve documents relevant to the occupation. We then create evidence cards from within the source documents. These cards serve as the standard against which automated systems can verify information. We test the models on both open-ended answers against the evidence cards and multiple choice questions using the same source material. We then go through a series of automated filters to ensure the quality and correctness of the questions. We also included a human evaluation step in which two annotators manually checked the quality of items, their sources and their answer options against a quality standard. Agreement between our manual and automated ``good'' decisions is about 67\% on the items. We use this as a metric to assess the performance of a scalable automated filter relative to expert curation. Correctness is defined in relation to the exact entailment in the source documents. All selected items are also subject to the same manually determined standard. We have created a curated authoritative ORQA dataset consisting of 480 questions, drawn from 187 source hosts, 116 occupations and all 21 major groups in the SOC. We test fifteen frontier and open-weight models without retrieval or browsing capability. We primarily test on closed-book multiple choice using three random seeds per question. We also test on an open-ended task where the model only sees the question text and must provide a brief free response which is scored by a three-judge panel. We also provide a wage-bill weighted measure that uses weighted occupations based on employment multiplied by the mean wage.

We observe three main results. One is that ORQA provides some separation between models in both the multiple choice and open- ended tasks. For multiple choice closed book tasks the frontier models (Claude Opus 4.6, GPT-5.4, and Claude Sonnet 4.6) are very close at 62.4\%, 60.3\% and 58.7\% respectively. Gemini 3.1 Pro is at 56.2\% and o3 at 51.7\%. The mid-tier models (Gemini 2.5 Flash and Claude Haiku 4.5) are in the 47-48\% range and the smaller or older models are 33-41\%. The absolute scores are lower than on general knowledge tasks because the questions are hand curated to be non-trivial and human verified, so what remains are questions that are genuinely difficult for the models. For open-ended tasks the absolute scores are again somewhat lower than closed book multiple choice, as expected. However the frontier ordering remains: GPT-5.4 at 54.2\%, Claude Opus 4.6 at 53.1\%, and Claude Sonnet 4.6 at 50.6\%. Wage-bill weighting does not affect the overall conclusion. The frontier models again appear first and the absolute scores again show only modest differences.

Second, aggregate scores mask significant occupation by model differences. For some occupations many models perform well while for others even the frontier models perform poorly. For instance GPT-5.4 performs overall at 60.3\% but performs at 0\% for Actuaries, Sheet Metal Workers and Fish and Game Wardens and at 11\% for Flight Attendants. We observe large cross-model differences (up to 100 points) in certain jobs. For Financial Managers, GPT-5.4 performs perfectly on all seeds and GPT-3.5 Turbo performs perfectly on none. Even at the level of high-level SOC categories, there is observed heterogeneity. For the most well-represented category, Healthcare Practitioners frontier models perform around 78-80\%, while for Installation, Maintenance, and Repair frontier models only perform 44-46\% and for Office and Administrative Support jobs frontier models perform 36-43\%. This is useful as it allows employers and employees to look beyond overall performance of a model and instead examine the reliability of a certain model for a certain type of job.

Third, ORQA observes a different entity but also tracks externally available signals. ORQA shows high correlations (Spearman) with GDPval (+0.90) and the GDPval-AA Elo ranking (+0.92) across overlapping models. ORQA also shows coverage across occupations in bins of exposure to the Anthropic Economic Index. This shows that ORQA both covers occupations that are both frequently and infrequently exposed to the Anthropic Economic Index and also tracks externally available signals. This implies ORQA does not currently appear to contradict externally available signals related to the economy and work. ORQA also adds the ability to perform source-aware, occupation-aware, and scalable model comparisons.

Finally, we present four contributions. Our first contribution is a scalable system for creating occupation-level knowledge resources from authoritative data. We present a general approach for transforming public data into verifiable questions and quality/balance considerations that are important to this task. Our second contribution is to instantiate this system as ORQA, a 480-item resource that contains data for 116 occupations, all 21 major groups in the SOC, and both digital and non-digital occupations. Our third contribution is a fifteen-model closed-book multiple choice and open response experiment, with uncertainty via cluster- bootstrap and an aggregation weighted by occupational wage bill. We observe significant variation across models, occupations, and SOC major groups. Our fourth contribution is to compare ORQA to external signals related to occupations such as GDPval, GDPval-AA, and exposure to the Anthropic Economic Index. We also demonstrate how the occupation-indexed system can be extended outside of authoritative data using a community-thread version that leverages data from public discussions (e.g. Reddit threads about occupations). ORQA is thus both a resource and an example of how far automation using data from specific sources can be extended to occupation-level understanding.

\section{Related Work}

\noindent\textbf{General capability metrics.}
Benchmarks for general purpose language models have been instrumental in tracking advancements in coding, scientific knowledge, reasoning, language understanding and other domains \citep{hendrycks2021mmlu,srivastava2023bigbench,liang2023helm,rein2024gpqa,chollet2019measure,jimenez2024swebench}. While useful for assessing capabilities, such benchmarks do not lend themselves to assessing impact on work or the economy because they are not structured around jobs. For instance, scoring high on an abstract capability benchmark does not necessarily imply that the model would assist a mechanic, groundskeeper, accountant, driver, or nurse with the concrete questions that occur in their respective fields.

Rather, evidence in the economics literature regarding the impact of AI on the job market focuses on exposure, usage, and field. Many economics papers have looked at exposure in order to identify the jobs or tasks that are most at risk from AI or generative AI \citep{brynjolfsson2018machines,webb2020impact,felten2021aioe,felten2023generative,eloundou2023gpts}. There have also been recent studies that examine usage \citep{handa2025economictasks,appel2025anthropicgeographic,chatterji2025chatgptusage,bick2024rapid}. Finally, field studies offer more concrete insights into quality and productivity impacts across specific domains like knowledge and customer service work \citep{brynjolfsson2025genaiwork,dellacqua2026jagged}.

There are several types of benchmarks that are relevant to understanding the economic impact of AI. Unlike ORQA, however, these do not typically provide a reusable test that evaluates many models on the same occupation-indexed questions. Instead, they generally evaluate on more realistic enterprise, agent, or work tasks. For example, recent benchmarks have transitioned from abstract skills to more realistic work. WebArena, WorkArena, WorkArena++, TheAgentCompany, and SWE-bench all test on software problems, enterprise processes, consequential workplace-like tasks, or web tasks for agents or models \citep{zhou2023webarena,drouin2024workarena,boisvert2024workarenaplus,xu2025agentcompany,jimenez2024swebench}. This represents a move toward more realistic evaluation. Still, these benchmarks tend to focus on digital, computer-mediated work. Recent efforts to connect agent benchmarks to occupations have revealed a significant bias toward computer and math-based occupations in benchmarks and a large underrepresentation of portions of the job market \citep{wang2026agentwork,wang2025humanworkflows}.

\noindent\textbf{Knowledge- and economically valuable benchmarks.}
GDPval is most similar in nature to ORQA. GDPval tests models on economically valuable occupation-based tasks. The tasks are created by experts and evaluated either by experts or models \citep{openai2025gdpval}. The benefits of GDPval are its realistic nature. However, the tasks are created by experts which makes scaling to hundreds of occupations repeatedly expensive. ORQA, on the other hand, scales by creating verifiable sources for occupation tasks using authoritative occupation information. Another effort that has begun to connect AI systems to ONET tasks highlights the importance of human-AI collaboration and human-AI assistance in model evaluation rather than solely end-to-end task solving \citep{shao2024cogym,chang2025chatbench}. ORQA is not in opposition to this effort, but rather evaluates a more narrow and specific aspect of this. Instead of measuring end-to-end task performance or the value of assistance, ORQA evaluates whether a model's performance is consistent with the authoritative knowledge base of an occupation.

\section{Methodology}\label{sec:method}

\paragraph{From exploratory to an approved pipeline.}
We created the current pipeline in two stages. In the first stage, we created an agentic pipeline with multiple automated steps that would take occupational data and from credible online sources and produce a large number of potential question-answer pairs. The objective of this initial stage was not to create the question bank itself, but to gain insight into what constitutes a good occupational question. A sample of these generated questions were manually inspected by two annotators. For each question, the answer choices, source, and question were scored using the same quality criteria. The annotators would then provide explanations for why a particular question- answer pair should be rejected. A tagging dashboard showed up several common types of problematic questions: answers not sufficiently implied by the source, distractors that could be answered using general domain knowledge, use of research articles instead of credible sources and correct answers obscured by vague modifiers. There is about a 67\% match between human and automated ``good'' annotations. We present this agreement as an indication of the similarity between automated scalable gates and expert curation, but do not present automation as a replacement for expert curation. Correctness itself is anchored to verbatim entailment from the source and is independent of this scalable gate. In the following, we describe the resulting system in its current state. The system produces items all of which have been filtered through the same quality validation step. We do not make distinctions between early and later items and do not keep track of the number of candidates rejected at each step. The overall pipeline from occupational sampling to a balanced pool is summarized in Figure~\ref{fig:method}.

\paragraph{Obtaining occupations.}
We start with the May 2024 national table from the Bureau of Labor Statistics Occupational Employment and Wage Statistics. For each Standard Occupational Classification (SOC) code, we calculate the wage bill as the sum of total employment nationwide times the average annual wage in that occupation. This represents the annual amount of labor compensation transiting through the occupation. We then allocate items across the major groups in the SOC according to the proportion of the nation's wage bill attributable to that group. This provides economic significance to both broad and narrow occupational categories. We try to get items for jobs in order of decreasing wage bill within each occupational category.

\paragraph{Domain whitelists per job.}
If a job had a general search on the web would yield millions of off-topic or poor quality pages such as content farms, marketing pages or blogs. Instead, we limit the search for sources for a given job to a list of trusted publishers curated from the official source information within O*NET. For example, a Registered Nurse may be sourced from domains like aacn.org (American Association of Critical-Care Nurses), nursingworld.org (American Nurses Association), ncsbn.org (National Council of State Boards of Nursing), state boards of nursing for that state, and the federal government sources bls.gov, osha.gov and cdc.gov. Each job has a whitelist appropriate to that job. These whitelists are complete. No document can come into the bank from a domain not on the whitelist. There is no overall allowlist from which academic or other documents can bypass the job specific whitelist.

\paragraph{Research papers are filtered at the source, not the gate.}
A research paper is not something that a professional is accountable for. Instead, it reports a result from a study. Because of this, any object that uses such a source fails in our grading scheme as a research\_paper\_source and is not eligible for a good grade. Such an object is essentially useless regardless of quality, and these sources are not used at all by the pipeline (no academic publishers are whitelisted by any occupation and no academic search is performed). Eliminating these sources before fetch, rather than rejecting them after generation, is safer (a paper that never enters the pipeline cannot survive a grader's mistake) and more cost effective (no extraction, generation, grading or fetch budget is wasted on objects that are guaranteed to be rejected). This distinction is made by document type and not by hosting source. Third party research is often archived by authoritative institutions. For example, a conference paper hosted on a .gov domain is still considered a research paper.

\paragraph{Source discovery and fetching.}
Discovery is a single general web round performed through a search API constrained to the whitelist for that job, so every candidate document comes from a publisher that governs the occupation. Nothing is pulled from outside the allowlist.

\paragraph{Evidence card and content creation.}
Documents are extracted and then parsed into structured objects. Documents are chunked with two and half to four kilobyte windows, overlapping a hundred tokens to avoid context drops. Each window is then passed through an evidence card extraction model (GPT-4o). Each card will extract one to three contiguous sentences from the document used as a source quote. The card will also extract the task or procedure referenced in the source quote, and the appropriate occupation.

\paragraph{Item creation.}
Each card is transformed into a six choice multiple choice question. The correct answer must be entailed by the source quote and can be created from the source quote. Distractor answers are created from within the same document. All of the above and None of the above are guaranteed to appear in every question as two of the six answer choices. Only one answer choice out of the six can be correct. Answer choices are also required to be of similar lengths and free from obvious hinting answers. If an item fails on either of these criteria it is recorded and the card is rejected.

\paragraph{Automated quality and categorization.}
After an item is created it goes through a series of automated filters. Some of these filters were created to prevent the failure modes discovered through manual testing. A failed item is sent to a specific regeneration attempt instead of being rejected immediately. The item that is produced by the regeneration attempt is then inserted back into the pipeline at the point of failure. Checks are performed in the following order: correct response is fully entailed by the original quote; question is complete and makes sense; all answer choices are topic-appropriate and not obviously nonsensical; no two responses are paraphrases of each other; only one response is correct based on the original; responses have a similar length and do not allude to use of specific sources (e.g., According to NFPA, . . .); the item is not solvable by all three of GPT-4o mini, Claude Haiku 4.5, and Gemini 2.5 Flash in parallel on a closed-book basis (and if so, is considered too easy and rejected); no answer choice can be eliminated based on general background knowledge (assessed by an independent LLM); and the correct answer does not camouflage behind vague terms such as standard or appropriate while other answers reference specific values or tools. The last filtering step is the one that is aligned to human annotations. It is this step that allows the pipeline to automatically reach the human defined ``good'' answer. Section~\ref{appdx:prompts} shows the exact prompts and the failure rate for each check.

\paragraph{Domain rebalance and occupation-floor rebalance.}
Once the quality checks are successful, domain balancing is performed across three domain rules to avoid over-representation from a single domain for any occupation. One rule ensures that no domain from a single source accounts for more than 30\% of items in that SOC. A second rule permits a larger percentage (40-50\%) for federal regulatory domains that are central to the occupation, and holds tangential domains to a smaller percentage (15-20\%). A third rule blocks a small number of low quality domains. Finally, an occupation-floor rebalance is performed which maintains all occupations with at least one verified item, while limiting the number of items in each occupation. This avoids situations in which a high-value occupation (such as registered nurses in cdc.gov) dominates the bank. There was a tendency toward concentration in available fields such as health care. To address this, we also performed a focused expansion for underrepresented jobs, with a minimum of three items in each job. This increased the number of jobs containing at least three items to 106 out of 116 jobs.

\paragraph{Final bank.}
The final ORQA bank consists of 480 items in 116 jobs across all 21 major SOC groups, from 187 distinct source hosts. About 43\% of the items come from federal government sources (ending in .gov). The rest come from licensing organizations and professional organizations, apart from two items from a state forestry extension service. osha.gov is the most prominent source host at 9.8\% of the bank, far below the 30\% maximum allowed from a source per job. The coverage is wide but not even: some major SOC groups (such as Production and Healthcare Practitioners) have many jobs, while others (such as Farming, Fishing, and Forestry) have only one or two jobs. We return to this issue in Section~\ref{sec:discussion}.

\paragraph{Evaluation.}
We evaluate fifteen models in closed-book testing. This means that the model is only shown the question text and the six answer choices. There is no browsing or retrieval. We test the following models: GPT-5.4, GPT-5.3-chat-latest, GPT-5.2, o3, GPT-4o, GPT-4o mini, GPT-3.5 Turbo, Claude Opus 4.6, Claude Sonnet 4.6, Claude Haiku 4.5, Gemini 3.1 Pro, Gemini 2.5 Flash, Llama 3.3 70B, DeepSeek V4 Pro, and Qwen 2.5 7B. Each of the 480 questions is run three times independently across each model using different random seeds. We report accuracy as the mean over all question-model-seed combinations. We estimate standard errors with a cluster bootstrap, where each cluster corresponds to an occupation. This is the correct level of clustering because items within the same occupation have similar content and cannot be assumed independent. For a multiple choice question with six possible answers, the accuracy by random guessing would be 1/6 $\approx$16.7\%. We include this as a reference line.

\section{Results}\label{sec:results}

\paragraph{Item validation.}
Prior to score interpretation we consider the degree to which ORQA items are occupational in nature. ORQA asserts three sources of validity. First, items are occupation-linked. Each item is associated with an O*NET occupation and derived from materials produced by industry associations, agencies, standards organizations, licensing organizations, regulatory organizations, academic institutions or other organizations that define or oversee the nature of work in that occupation. Second, response correctness is source-derived and not model-inferred. That is, the correct response must be entailed by an exact excerpt from the source material. Finally, items undergo the automated verifications detailed in Section~\ref{sec:method} that were tuned using human annotation before being added to the bank. The Veterinarians call-out in Figure~\ref{fig:method} is an example of this. An OSHA exposure guideline in the workplace specifies an absolute level for halogenated anesthetic agents. ORQA transforms this into an occupational item and the correct response is correct because it is present in the source and not because a model has inferred its plausibility.

\begin{figure}[!t]
  \centering
  \caption{Overall ORQA performance across multiple-choice, open-ended, and wage-weighted evaluation.}
  \includegraphics[width=\linewidth]{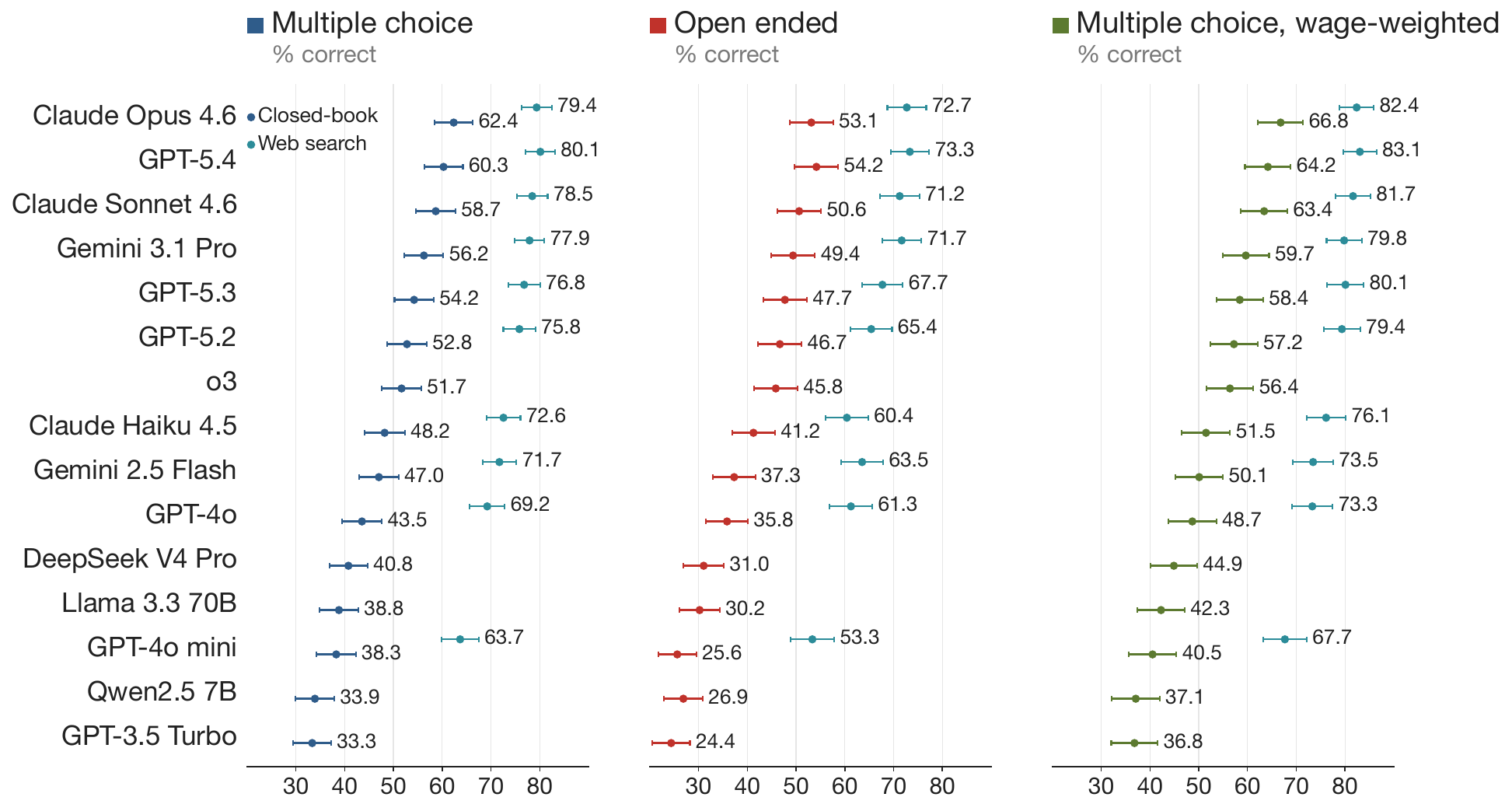}
  \figurenote{All panels evaluate the same occupation-indexed item bank and model slate. Each panel plots two points per model where available: a closed-book score with no tools and a web-search score for the ten models with a native web-search tool, while the five models without web search show the closed-book score only. Closed-book multiple-choice and wage-weighted scores average three seeds per item, open-ended scores use a three-judge ensemble, and wage weighting changes only the aggregation weights. Bars show cluster-bootstrap standard errors resampled at the occupation level.}
  \label{fig:main_performance}
\end{figure}

\paragraph{Overall performance.}
The same 480 item bank is assessed in three different ways: closed-book multiple choice, open-ended responses with model-scored responses, and wage-bill-weighted closed-book accuracy (Figure~\ref{fig:main_performance}). These three perspectives yield similarly comparable results. In closed book multiple choice Claude Opus 4.6 comes first at 62.4\%, then comes GPT-5.4 at 60.3\%, and then Claude Sonnet 4.6 at 58.7\%. The three frontier models are very close, within about 4 percentage points of each other. Then comes Gemini 3.1 Pro at 56.2\%, GPT-5.3-chat-latest at 54.2\%, GPT-5.2 at 52.8\%, and o3 at 51.7\%. The middle models like Gemini 2.5 Flash at 47.0\% and Claude Haiku 4.5 at 48.2\% are around half. GPT-4o is at 43.5\%. Open models range from 33.9\% to 40.8\%. And GPT-3.5 Turbo is at 33.3\%. All fifteen models perform above the random chance score of 16.7\% but due to the nature of the test bank (only containing non-trivial questions) even the top performing models still get over a third of the occupation-related questions wrong.

In open-ended output the scores are reduced but the frontier ranking is maintained. For open-ended output each model is presented with only the question text and no options or source material. Then the model generates a free form response in at most five lines to the question. To help mitigate the problem of a single judge being biased toward a particular model three judges from different providers are used. There is one OpenAI GPT judge, one Claude judge and one Gemini judge. Each judge is given the question, the correct answer and an anonymized answer from the model under evaluation. The final score for each model is a binary value determined by majority vote among the three judges. In this set, GPT-5.4 comes in first at 54.2\%, then Claude Opus 4.6 at 53.1\% and Claude Sonnet 4.6 at 50.6\%. As expected for answer generation versus recognition, the scores are about 7-9 points lower than on multiple choice. However, the order remains fairly similar. Having access to a basic web search tool greatly improves performance on both tasks. The frontier models perform at about 71-73\% on open-ended and 78-80\% on multiple choice, which is a significant improvement over the penalty from closed-book testing. The other five models without access to a web search tool are only scored in closed-book mode. The leaderboards are also fairly robust to economic weighting. Using the national wage bill for each occupation, Claude Opus 4.6 again comes in first at 66.8\%, then GPT-5.4 at 64.2\% and then Claude Sonnet 4.6 at 63.4\%. The similarity of the results across multiple choice, open-ended, and economic weighting implies that the aggregate ordering is not driven by the six options for each task or by the unweighted occupation mix.

\begin{figure}[!t]
  \centering
  \caption{SOC major-group heatmap of ORQA accuracy by model.}
  \includegraphics[width=\linewidth]{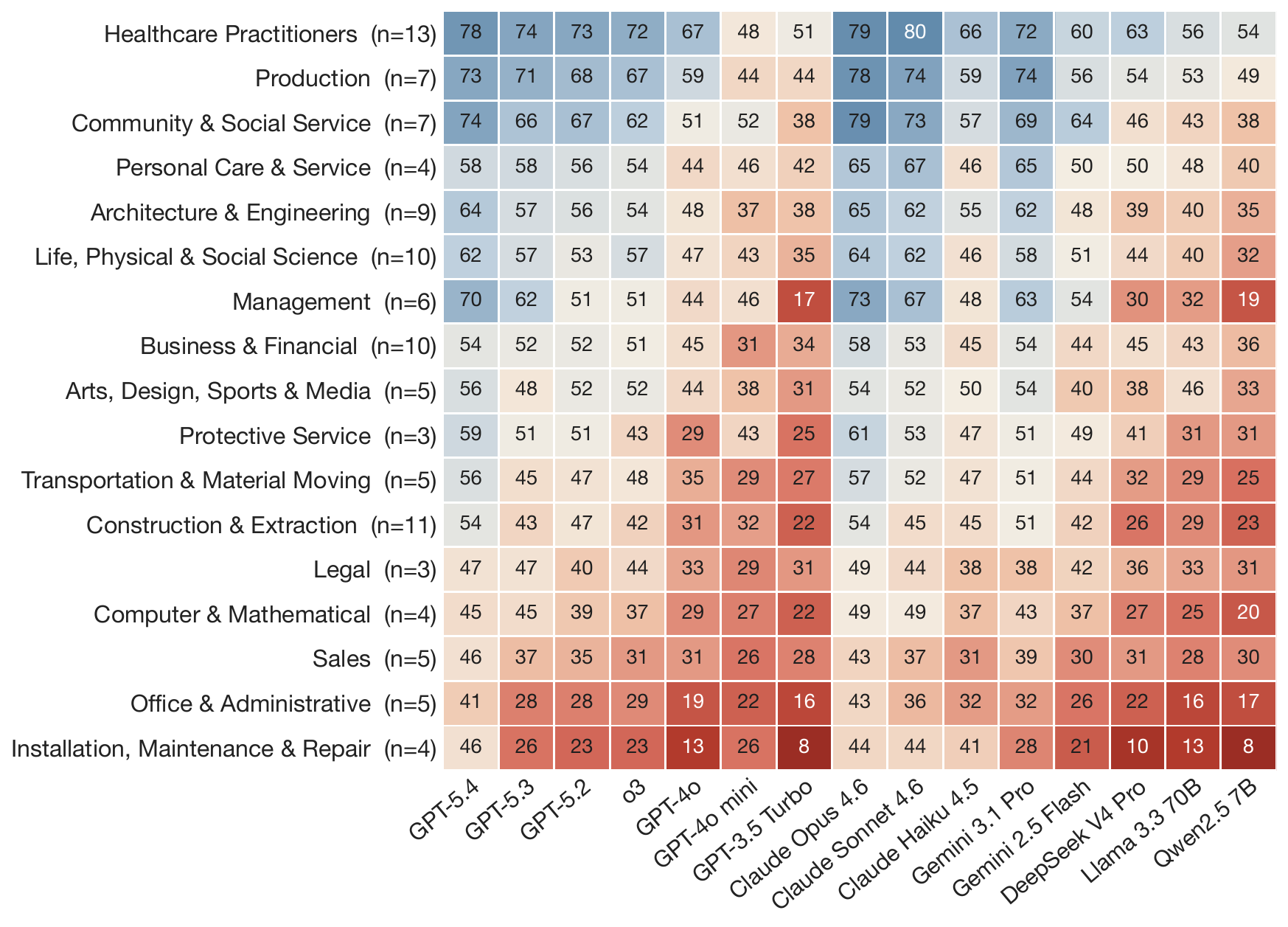}
  \figurenote{SOC rows aggregate item-level accuracy within broad BLS occupational groups. Each cell averages all retained items in the group across the model's three seeds. Groups vary widely in how many occupations they contain, and thinly covered groups should be interpreted with that in mind.}
  \label{fig:occ_heatmap_soc}
\end{figure}

\begin{figure}[!t]
  \centering
  \caption{Occupation-level heatmap showing model-specific strengths and failures.}
  \includegraphics[width=\linewidth]{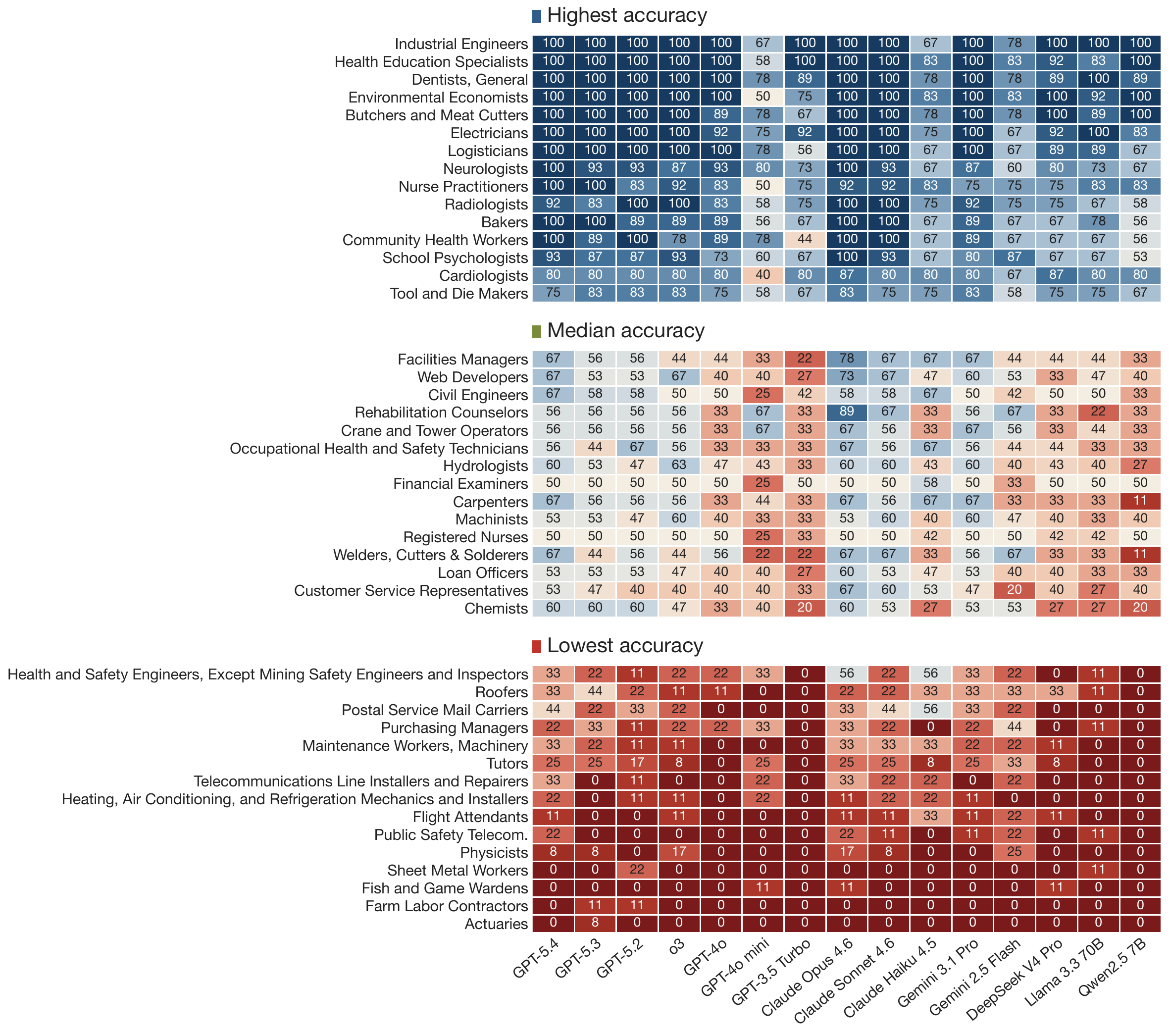}
  \figurenote{Occupations are ordered by cross-model mean accuracy so that easy and hard domains are visually separated. Cell values summarize model accuracy within an occupation and should be interpreted with item counts in mind.}
  \label{fig:occ_heatmap_quartile}
\end{figure}

\paragraph{Occupational heterogeneity.}
There is a great deal of heterogeneity across occupations that is masked by aggregate leaderboards. There is significant variation in both the overall performance level and the ranking across broad occupational categories at the SOC major-group level (Figure~\ref{fig:occ_heatmap_soc}). Frontier performs 78-80\% on the most comprehensively covered group, Healthcare Practitioners (13 occupations with 52 items), 44-46\% on Installation, Maintenance, and Repair and 36-43\% on Office and Administrative Support. We would like to emphasize, though, that we should not over interpret the group level results. There are a number of SOC major groups that are only covered by a few occupations in the current inventory. For example, as Table~\ref{tab:soc_groups} shows, Food Preparation and Serving contributes only one item and Farming, Fishing, and Forestry only one occupation; both are too thin to plot and are omitted from the heatmap. Group level results for such sparse groups should be treated provisionally rather than as definitive. This is also seen more clearly at the occupation level in Figure~\ref{fig:occ_heatmap_quartile}. For occupations with three or more items, easily performed occupations like Electricians, Dentists, and Industrial Engineers appear in the 90\%+ range across many models. However, there are also a number of occupations that are near zero in frontier models. GPT-5.4 gets 60.3\% overall but gets no seed right for Actuaries, Sheet Metal Workers, or Fish and Game Wardens or 11\% for Flight Attendants.

There are also very large cross-model discrepancies. For Financial Managers, GPT-5.4 gets all the seeds right and GPT-3.5 Turbo gets none. These are also the examples where domain-specific discrepancies occur and leaderboard aggregates will not be able to capture.

\paragraph{What difficult items look like.}
One of the benefits of ORQA is that it can access to jobs that are not typically included in job assessments because they are in the physical or non-digital world. Also, difficult items can be concrete instead of abstract. There are a few trends that can be pointed out. One is a failure to accurately recall specific codes and numeric values. For example, in an OSHA item regarding Slaughterers and Meat Packers, a specific exposure limit is referenced when a piece of equipment makes impact or impulse noises. The models do not appear to do well in recalling these specific values in closed-book. Another example is in an item for Tool and Die Makers which references a specific machine safety document. The question requires what must be checked prior to inching the ram on a punch press. The correct answer is a specific action rather than something like follow all safety procedures. Second, many failures are retrieval-shaped. The correct response to a question is some piece of fact from some authoritative source that a web-search-capable model can retrieve, but not a closed-book model. An example from Flight Attendants uses a question from GoJet's collective bargaining agreement regarding the boarding pay for flights that are delayed and subsequently deplaned and reboarded. A closed-book model performs very poorly on this question, but a web-search capable model performs well because it can retrieve the collective bargaining agreement. These failures illustrate both the kind of knowledge that ORQA evaluates (professionally relevant knowledge that is not currently in frontier models) and also the possibility for a large gap between closed-book performance and web-search performance.

\paragraph{Validation.}
We are fairly conservative about our claims to validation. Our strongest claim is that ORQA's rankings are not contradicted by independent external evidence. Because there is currently no authoritative-source indexed occupation benchmark available, we compare to three external metrics rather than expecting exact agreement with any one of them. First, GDPval reports long form performance against professional experts in their respective occupations. There is strong correlation between ORQA performance and GDPval win rate (Spearman $\rho$ = +0.90, Pearson r = +0.89) across the five models that overlap our slate. Second, GDPval-AA, the Artificial Analysis version of GDPval \citep{artificialanalysis2025gdpvalaa}, reports a blind pairwise Elo on long form economic tasks. Across the eleven models that overlap our slate, Spearman $\rho$ = +0.92 and Pearson r = +0.93. On both leaderboards, the frontier of GPT-5.4, Claude Sonnet 4.6 and Claude Opus 4.6 are at the top. The open-weight models are in the lower positions on both leaderboards, and the relative ordering in between is largely maintained. Third, we compare the coverage of ORQA to the Anthropic Economic Index \citep{anthropic2025economicindex}. We group jobs based on exposure to observed usage of Claude. ORQA covers jobs that are either highly used, or low used jobs that are not well-represented in many usage-based measures of AI in the workplace by absence in those jobs. We believe these evaluation methods are rather basic, and more in-depth evaluation against human-compiled gold answers across jobs would be advantageous.

\section{Discussion}\label{sec:discussion}

ORQA is intended to be a method for creating a source-grounded agreement to professional advice/guidelines, rather than a full representation of occupational AI ability. The key advantage of ORQA is enabling scalability to the level of occupation. Many occupations have authoritative materials that outline safety, professional standards, reporting responsibilities, thresholds, procedures and other aspects. These can be transformed into testable questions. This enables comparisons over a significantly larger portion of the occupation landscape compared to comparisons using only expert-created work sample tests. There are however, a number of constraints introduced by this approach.

Firstly, some limitations to the approach should be discussed. ORQA uses authoritative materials as ground truth. However, there are aspects of the profession that do not only exist in these materials. Due to the possible difference in relevance and incompleteness of the source materials, an accurate answer from ORQA should be seen as agreement to a specified authoritative or professional source rather than an indication that the model would make the correct decision in a professional context. Second, coverage is extensive but uneven, and the total coverage remains relatively limited: the current bank covers 116 occupations in all 21 SOC major groups, but the number of items per occupation and per some SOC major groups is quite low (some SOC major groups have only one or two occupations). Some occupations (such as production and healthcare occupations) generate many high-quality, sourceable items, while others generate only a few items. A small number of SOC major groups (such as Farming, Fishing, and Forestry) are also sparse because there are a limited number of authoritative machine-readable sources. Thus, estimates at the occupation level for occupations with few items and estimates at the SOC group level for thinly covered SOC groups should be viewed as exploratory rather than as definite rankings. We are also careful about statements of generalizability to the full economy: ORQA does cover a substantial and expanding share of the economy, but not all occupations and sectors in the economy. It would be useful to expand the bank to cover occupations and sectors that are under-covered and have fewer authoritative sources available.

Third, the use of our sources may pose a memorization problem since they may be present in pre-training data for models. We attempt to tackle the contamination issue empirically via a per-model pre training cutoff stratification included in Appendix~\ref{appdx:contamination}. For items taken from a passage published after a given model's training cutoff (thus being free from contamination for that model), performance is equal to or better than performance on pre-cutoff passages for 14 of the 15 models analyzed. This is in contrast to the performance one would expect under a pure memorization hypothesis. We also provide an alternative robustness test by analyzing performance on an open-ended task in addition to the multiple choice tasks. This shows that ranking is not solely a function of correct multiple choice answer selection. We can continue to investigate the contamination problem. ORQA also does not facilitate significant human-LLM back and forth or deep human collaboration. Thus we believe ORQA to be one element of occupational assessment. It provides a useful signal for occupational knowledge but should be used in conjunction with human in the loop, task based, and interactive assessments before use in practice. Our approach can also be used to evaluate end to end occupational tasks using AI agents in the future.

Our main contribution is a scalable approach to occupational evaluation. ORQA provides a way to connect occupational knowledge to authoritative online resources and transform those resources into source verifiable evaluation tasks at a significantly lower cost than expert based evaluation. The same approach can also be used for non-authoritative resources. As a complementary effort, we also explore a community-thread based approach that connects jobs to Reddit threads (Appendix~\ref{appdx:community}). Several of the issues mentioned above can lend themselves quite naturally to future research instead of being inherent limitations to the approach. A greater number of models and compute will allow for exploring more sources, producing more candidate items per job, testing more models, and performing more in-depth robustness tests. Greater human curation will allow for improved alignment between source quality and practicality in the real world. We believe this approach provides a scalable foundation for progressively more realistic job-level evaluations, rather than being the ultimate solution for assessing performance in the real world.

\bibliographystyle{plainnat}
\bibliography{references}

\FloatBarrier

\appendix

\section{Technical Appendices and Supplementary Material}

We structure our appendices around diagnostics for the question are good argument and also on a community-thread version of ORQA that complements the main implementation. Appendix~\ref{appdx:bank} contains information about bank composition, occupation and source lists, complete source distributions and SOC group information. Appendix~\ref{appdx:diag} contains per-item diagnostics including pairwise model significance, per-item difficulty distributions, inter-model correctness correlation, and the release date trend across models. Appendix~\ref{appdx:contamination} contains a diagnosis of contamination during pre- training by comparing pre- and post-cutoff accuracy strata. Appendix~\ref{appdx:prompts} contains details about the prompts, list of models used, and inference parameters. Appendix~\ref{appdx:validation} contains information about external validation using three externally available signals: GDPval, GDPval-AA Elo, and Anthropic Economic Index exposure. Appendix~\ref{appdx:dashboard} contains information about the interactive dashboard. Appendix~\ref{appdx:heatmap} contains the complete heatmap across occupations. Appendix~\ref{appdx:community} contains the community-thread version of ORQA.

\subsection{Composition of Bank and Lists}\label{appdx:bank}

\paragraph{Number of items and occupations}
All occupations in the ORQA bank and the number of bank items in each occupation are listed in Table~\ref{tab:occupations} in the appendix. Each occupation also has an O*NET SOC code. The final bank consists of 480 items across 116 occupations across all 21 major groups in the SOC. Following the addition of occupation- floor items, 106 of the 116 occupations have three or more items while the other occupations have either one or two items.

\paragraph{Sources: URL, domain and category}
Every authoritative source URL that contributed at least one item to the bank is listed in Table~\ref{tab:sources} in the appendix. For each source, the domain, category and URL is listed. Table~\ref{tab:source_mix} shows the total distribution of source categories and the most commonly used source hosts. 57.1\% of the items come from licensing boards, associations and professional societies. 42.5\% of the items come from federal and state government hosts (.gov). Only 0.4\% of the items come from educational institutions or .edu hosts. This lack of educational sources is not an emergent characteristic of the bank. Rather, it results from the source filtering explained in Section~\ref{sec:method}. Academic search is never performed and no research article is ever retrieved because research publishers are not on any whitelist for any occupation. There are zero items in the bank from a research source, and 0.4\% of the bank consists of two items from tfsweb.tamu.edu, the Texas A\&M Forest Service. The Texas A\&M Forest Service is an extension of the state forestry program and publishes practice standards for forestry, not a research organization. The bank consists of 187 different source hosts. The most commonly occurring host in the bank is osha.gov, which accounts for 9.8\% of the bank.

\subsection{Item and Model Diagnostics}\label{appdx:diag}

\paragraph{Correct option position distribution.}
Proportion of items with correct answer in position A, B, C, D, E, or F. For items A through D, the order of the substantive answer selections is randomized across items. For items E and F, the answer selections are always All of the above and None of the above respectively. We also provide accuracy per model for cases where the correct answer is a non-substantive answer and for cases where the correct answer is a substantive answer.

\paragraph{Pairwise comparisons between models.}
We perform pairwise comparisons between the fifteen models using cluster bootstrapping on the level of occupation across 3 observations per model and 480 items per model. From these comparisons we provide a pairwise significance matrix that can be used by users to assess the relative significance of pairwise orderings in the leaderboard. Especially interesting are the three comparisons between the top cluster of Claude Opus 4.6, GPT-5.4, and Claude Sonnet 4.6, and the middle comparison between GPT-5.3-chat-latest and Gemini 3.1 Pro.

\begin{figure}[!htbp]
  \centering
  \caption{The ORQA item bank spans easy, hard, and discriminating middle-difficulty questions.}
  \label{fig:itemdiff}
  \includegraphics[width=0.85\linewidth]{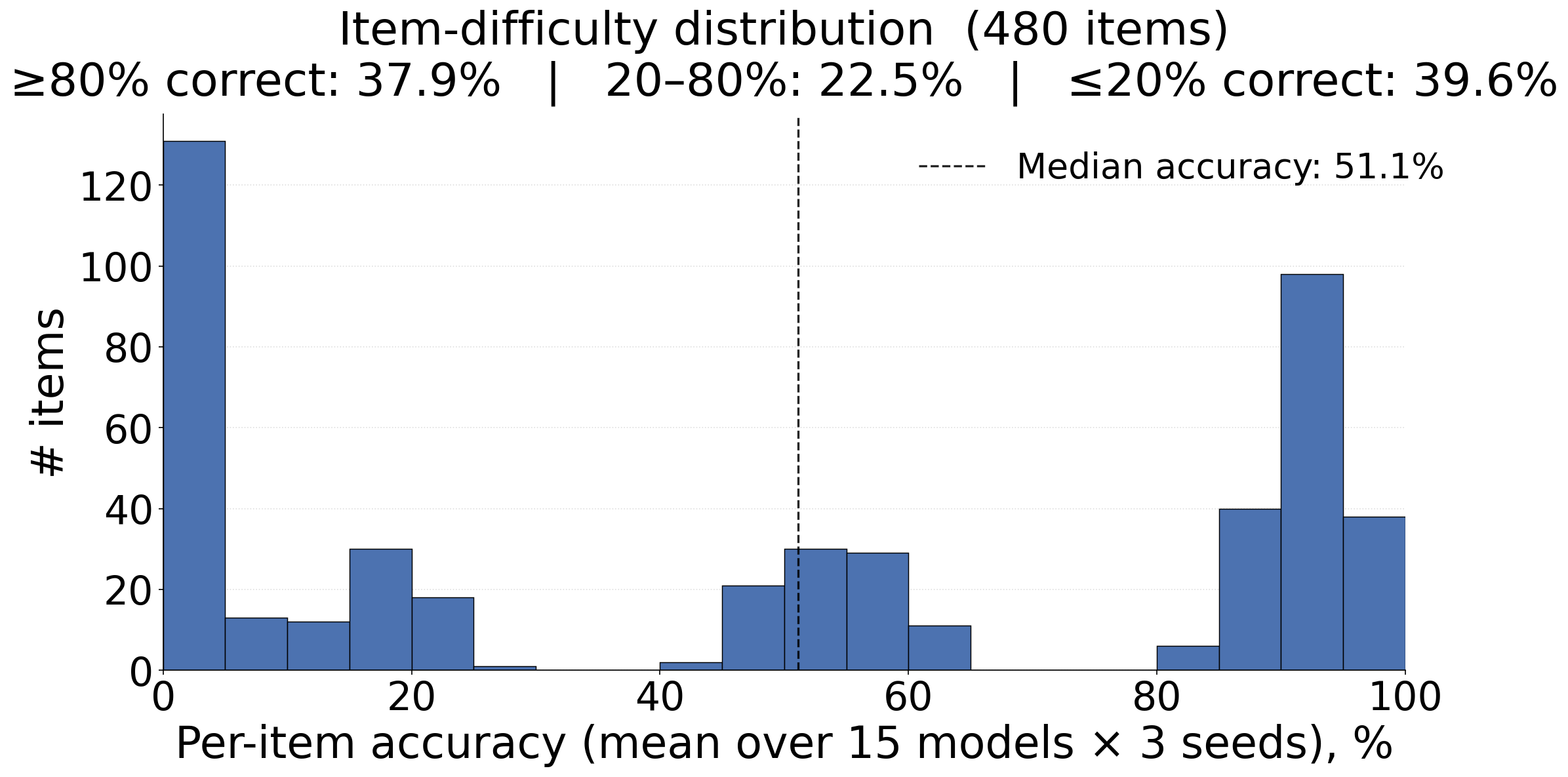}
  \figurenote{Each item contributes one mean accuracy over 45 model-seed observations. The curated bank is deliberately weighted toward the discriminating and hard regions of the difficulty distribution.}
\end{figure}

\paragraph{Item-difficulty distribution.}
Figure~\ref{fig:itemdiff} displays the distribution of per-item accuracy across fifteen models and three random seeds. We observe a bimodal distribution in which about 38\% of items is correct with at least 80\% accuracy across all models, 22\% straddle the discriminative middle region (between 20\% and 80\% correct), and 40\% are correct with no more than 20\% accuracy. The median per-item accuracy is 51.1\%. The human-curated set has a significantly higher relative proportion of truly difficult items than the pre-curated set. This explains the heavy lower mode.

\paragraph{Per-occupation difference between models.}
Table~\ref{tab:discrepancy} displays some instances of inter-occupation differences that are obscured by the aggregate leaderboards. We divide the table into three sections. Section (a) shows occupations in which the difference between the best and worst model is at least 75 percentage points across models. Section (b) shows occupations in which GPT-5.4 achieves zero accuracy despite its overall accuracy of 60.3\%. This reveals that even the strongest evaluated model has specific blind spots. Section (c) shows occupations in which the small open-weight Qwen 2.5 7B model (33.9\% accuracy) performs the same as the frontier group. These are more relevant to a practitioner choosing a model from the set based on specific tasks rather than the global ordering.

\begin{table}[!htbp]
  \centering
  \caption{Per-occupation discrepancy across models on the ORQA bank. Each cell is per-occupation accuracy (\%) averaged over three seeds, restricted to occupations with at least three items. \emph{Spread} is the maximum minus minimum accuracy across all 15 evaluated models. \textbf{(a)} Largest cross-model spread. \textbf{(b)} GPT-5.4 answers no seed correctly despite its overall 60.3\%. \textbf{(c)} Qwen 2.5 7B (overall 33.9\%) matches the frontier.}
  \label{tab:discrepancy}
  \small
  \begin{tabular}{lrrrrr}
    \toprule
    Occupation & $n$ & GPT-5.4 & Opus 4.6 & Sonnet 4.6 & Qwen 2.5 7B \\
    \midrule
    \multicolumn{6}{l}{\textit{(a) Frontier models disagree (largest cross-model spread)}} \\
    Financial Managers                   & 3  & 100 & 100 & 89  & 0  \\
    Intelligence Analysts                & 3  & 100 & 100 & 78  & 11 \\
    Aerospace Engineers                  & 3  & 78  & 89  & 78  & 22 \\
    Healthcare Social Workers            & 4  & 75  & 83  & 83  & 8  \\
    \cmidrule(lr){1-6}
    \multicolumn{6}{l}{\textit{(b) GPT-5.4 fails entirely (entire frontier weak)}} \\
    Actuaries                            & 4  & 0   & 0   & 0   & 0  \\
    Fish and Game Wardens                & 3  & 0   & 11  & 0   & 0  \\
    Sheet Metal Workers                  & 3  & 0   & 0   & 0   & 0  \\
    Physicists                           & 4  & 8   & 17  & 8   & 0  \\
    \cmidrule(lr){1-6}
    \multicolumn{6}{l}{\textit{(c) Open-weight Qwen 2.5 7B matches the frontier}} \\
    Environmental Economists             & 4  & 100 & 100 & 100 & 100 \\
    Butchers and Meat Cutters            & 3  & 100 & 100 & 100 & 100 \\
    Industrial Engineers                 & 3  & 100 & 100 & 100 & 100 \\
    Cardiologists                        & 5  & 80  & 87  & 80  & 80  \\
    \bottomrule
  \end{tabular}
\end{table}

\begin{figure}[!htbp]
  \centering
  \caption{Per-item correctness correlations show which models share the same ORQA successes and failures.}
  \label{fig:correl}
  \includegraphics[width=0.7\linewidth]{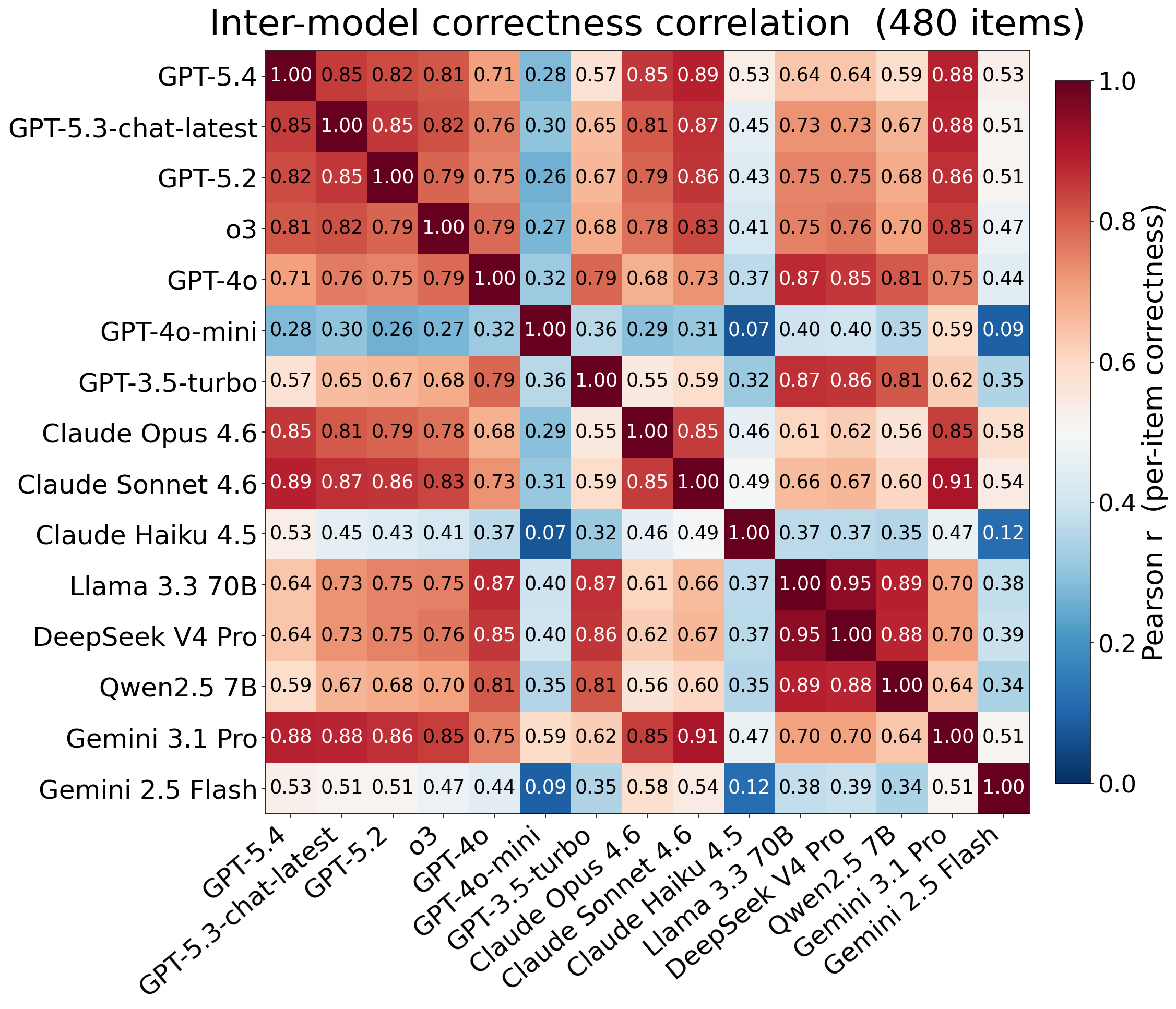}
  \figurenote{Correlations use binary per-item correctness vectors rather than aggregate scores. Correlations involving the three Pass-7 probe models are attenuated by construction.}
\end{figure}

\paragraph{Inter-model correlation.}
Figure~\ref{fig:correl} shows the Pearson correlation between models on item-level correctness in the ORQA bank. Across this curated benchmark, the higher performing models all correlate highly with each other across provider families. For example, GPT-5.4 correlates r = 0.94 with Claude Opus 4.6, Claude Opus 4.6 correlates r = 0.93 with Sonnet 4.6 and the open-weight models (Qwen 2.5 7B, DeepSeek V4 Pro and Llama 3.3 70B) all correlate between r = 0.91 and r = 0.92. This makes sense because the ORQA bank appears to test along a professional-knowledge axis. The low correlations all involve GPT-4o mini, Gemini 2.5 Flash and Claude Haiku 4.5. The GPT-4o mini and Claude Haiku 4.5 have the lowest correlation (r = 0.07). For the closed-book difficulty assessment (Pass 7), these three models were used. During bank creation, responses that were correct across all three models were removed. This causes an automatic reduction and in some cases even reversal of the pairwise correctness relationships between these models. Such pairwise relationships should be analyzed with this in mind.

\begin{figure}[!htbp]
  \centering
  \caption{ORQA accuracy improves across approximate model release dates within provider families.}
  \label{fig:release}
  \includegraphics[width=0.95\linewidth]{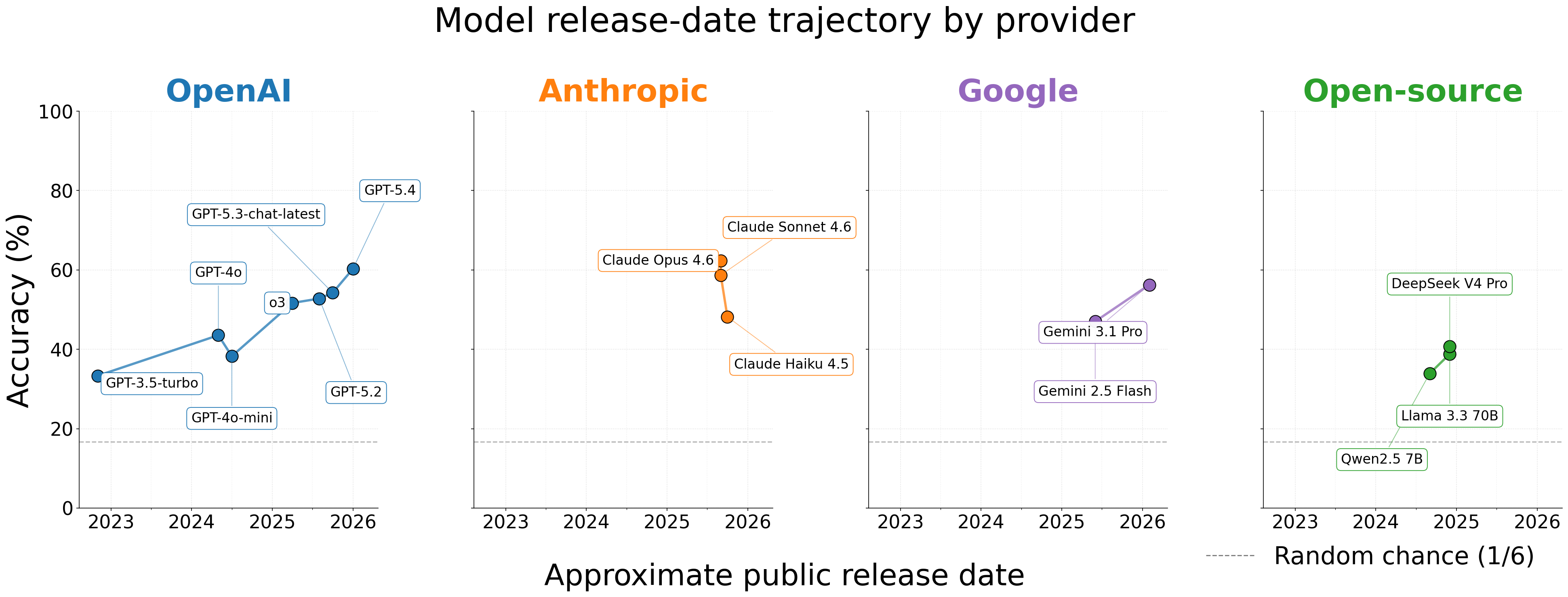}
  \figurenote{Public release date is a coarse proxy for model generation, not a causal variable. The dashed line marks six-option random chance.}
\end{figure}

\paragraph{Release-date trends.}
Figure~\ref{fig:release} displays the accuracy across each model by the estimated date of public release, separately for each provider. For OpenAI, Anthropic and Google, across each of their product lines, accuracy increases with date of release. The open-weight models also show an increase with release date, though weaker. Frontier model performance increases by about 27 percentage points from GPT-3.5 Turbo (33.3\%) to GPT-5.4 (60.3\%). We note, however, that date of release is only a noisy indicator of effort put into performance and some of the performance increase might be due to more focused post-training on professional material.

\paragraph{Source distribution.}
The distribution of sources and the top hosting sources is shown in Table~\ref{tab:source_mix}. Professional licensing and associations, boards and societies account for 57.1\% of the content, state and federal government sources account for 42.5\%, and .edu sources account for only 0.4\%, two items from a state forestry extension service rather than a research venue. The largest host is osha.gov at 9.8\% of the bank, significantly under the 30\% limit on the source-mix per occupation.

\begin{table}[!htbp]
  \centering
  \caption{Source distribution of the ORQA bank (480 items, 116 occupations, 187 unique source hosts). Top: source category mix as a share of items. Bottom: top ten source hosts by item count.}
  \label{tab:source_mix}
  \small
  \begin{tabular}{lr}
    \toprule
    Source category & Share of items \\
    \midrule
    Professional societies, associations \& licensing boards & 57.1\% \\
    Federal \& state government (\texttt{.gov}) & 42.5\% \\
    Academic / \texttt{.edu} & 0.4\% \\
    \midrule
    \multicolumn{2}{l}{\textit{Top source hosts by item count}} \\
    \texttt{osha.gov} & 47 (9.8\%) \\
    \texttt{stacks.cdc.gov} & 26 (5.4\%) \\
    \texttt{cdc.gov} & 21 (4.4\%) \\
    \texttt{epa.gov} & 12 (2.5\%) \\
    \texttt{faa.gov} & 10 (2.1\%) \\
    \texttt{docinfofiles.nfpa.org} & 8 (1.7\%) \\
    \texttt{amftrb.org} & 7 (1.5\%) \\
    \texttt{ecfr.gov} & 6 (1.2\%) \\
    \texttt{ashrae.org} & 6 (1.2\%) \\
    \texttt{fjc.gov} & 6 (1.2\%) \\
    \bottomrule
  \end{tabular}
\end{table}

\paragraph{Composition within SOC major groups.}
Table~\ref{tab:soc_groups} shows the number of occupations and items within each SOC major group. All 21 SOC major groups are included but there are significant differences in the number of items and occupations within each major group. Healthcare Practitioners, Life/Physical/Social Science and Business \& Financial have more than 40 items in each of ten or more occupations. However, Food Preparation \& Serving, Farming/Fishing/Forestry, Education \& Library and Building \& Grounds have only one or two occupations and a small number of items. For these sparse groups, the accuracy within the group should be considered somewhat tentative.

\begin{table}[!htbp]
  \centering
  \caption{SOC major-group composition of the ORQA bank, sorted by item count. Group-level results for thinly covered groups (bottom rows) should be interpreted with caution.}
  \label{tab:soc_groups}
  \small
  \begin{tabular}{llrr}
    \toprule
    SOC code & Major group & Items & Occupations \\
    \midrule
    29 & Healthcare Practitioners \& Technical & 52 & 13 \\
    19 & Life, Physical \& Social Science & 52 & 10 \\
    13 & Business \& Financial Operations & 44 & 10 \\
    17 & Architecture \& Engineering & 42 &  9 \\
    21 & Community \& Social Service & 32 &  7 \\
    47 & Construction \& Extraction & 31 & 11 \\
    51 & Production & 26 &  7 \\
    53 & Transportation \& Material Moving & 25 &  5 \\
    43 & Office \& Administrative Support & 23 &  5 \\
    11 & Management & 21 &  6 \\
    41 & Sales \& Related & 18 &  5 \\
    15 & Computer \& Math & 17 &  4 \\
    33 & Protective Service & 17 &  3 \\
    27 & Arts, Design, Entertainment, Sports \& Media & 16 &  5 \\
    39 & Personal Care \& Service & 16 &  4 \\
    23 & Legal & 15 &  3 \\
    49 & Installation, Maintenance \& Repair & 13 &  4 \\
    37 & Building \& Grounds Cleaning \& Maintenance & 9 &  1 \\
    25 & Educational Instruction \& Library & 7 &  2 \\
    45 & Farming, Fishing \& Forestry & 3 &  1 \\
    35 & Food Preparation \& Serving & 1 &  1 \\
    \midrule
    & Total & 480 & 116 \\
    \bottomrule
  \end{tabular}
\end{table}

\begin{figure}[!htbp]
  \centering
  \caption{Average ORQA accuracy is weakly related to occupational wage, and unrelated after employment weighting.}
  \label{fig:wage_scatter}
  \includegraphics[width=0.95\linewidth]{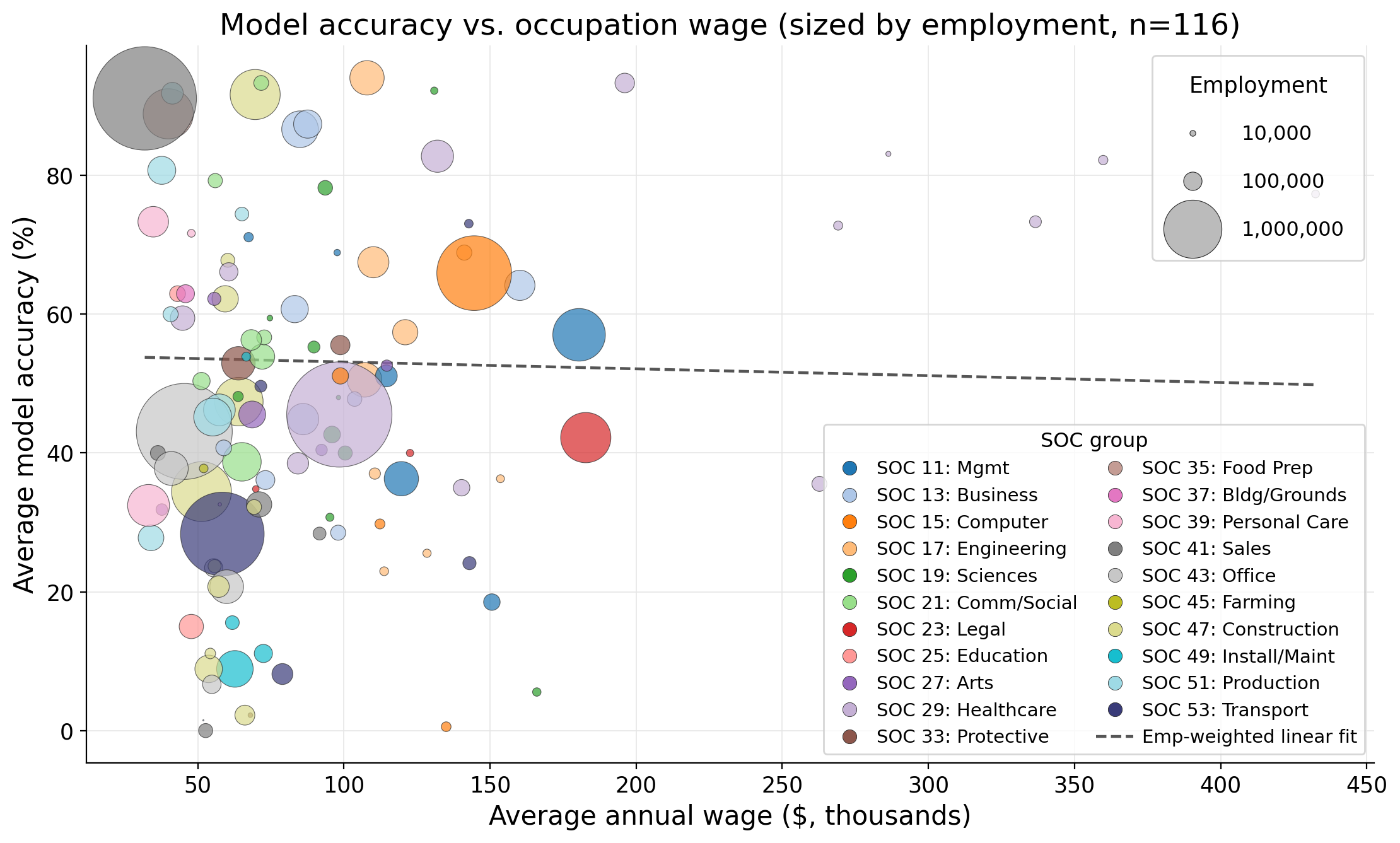}
  \figurenote{Marker size reflects national employment, so visually prominent points are large occupations. The dashed trend is the employment-weighted fit and is descriptive; it should be read alongside the broad dispersion at each wage level.}
\end{figure}

\paragraph{Model accuracy by occupation wage.}
Figure~\ref{fig:wage_scatter} displays the relationship between BLS OEWS 2024 annual mean wage and per-occupation model accuracy (average across fifteen models). Each point is scaled according to the total number of employment in the United States. Across all occupations the relationship is weak. The unweighted Pearson correlation between model accuracy and wage is r = +0.22 (p = 0.02). Over the entire range of wages there is also approximately the full range of model accuracy. However, when employment across the nation is used as a weight for occupations the relationship becomes almost non-existent (r = $-$0.02, with a weighted fit close to zero). We only perform a descriptive analysis. There does not appear to be a strong link between occupational wage and accuracy on procedural knowledge of professional tasks using closed book exams. There also does not appear to be a significant concentration of AI ability within the high-wage share of the labor market when weighted by employment. We present this as an inductive result using our data and do not attempt to test a specific hypothesis here. A more thorough analysis can be performed in future work.

\begin{figure}[!htbp]
  \centering
  \caption{Model accuracy on digital versus non-digital occupations, and the hardest occupations by category.}
  \label{fig:digital_nondigital}
  \includegraphics[width=\linewidth]{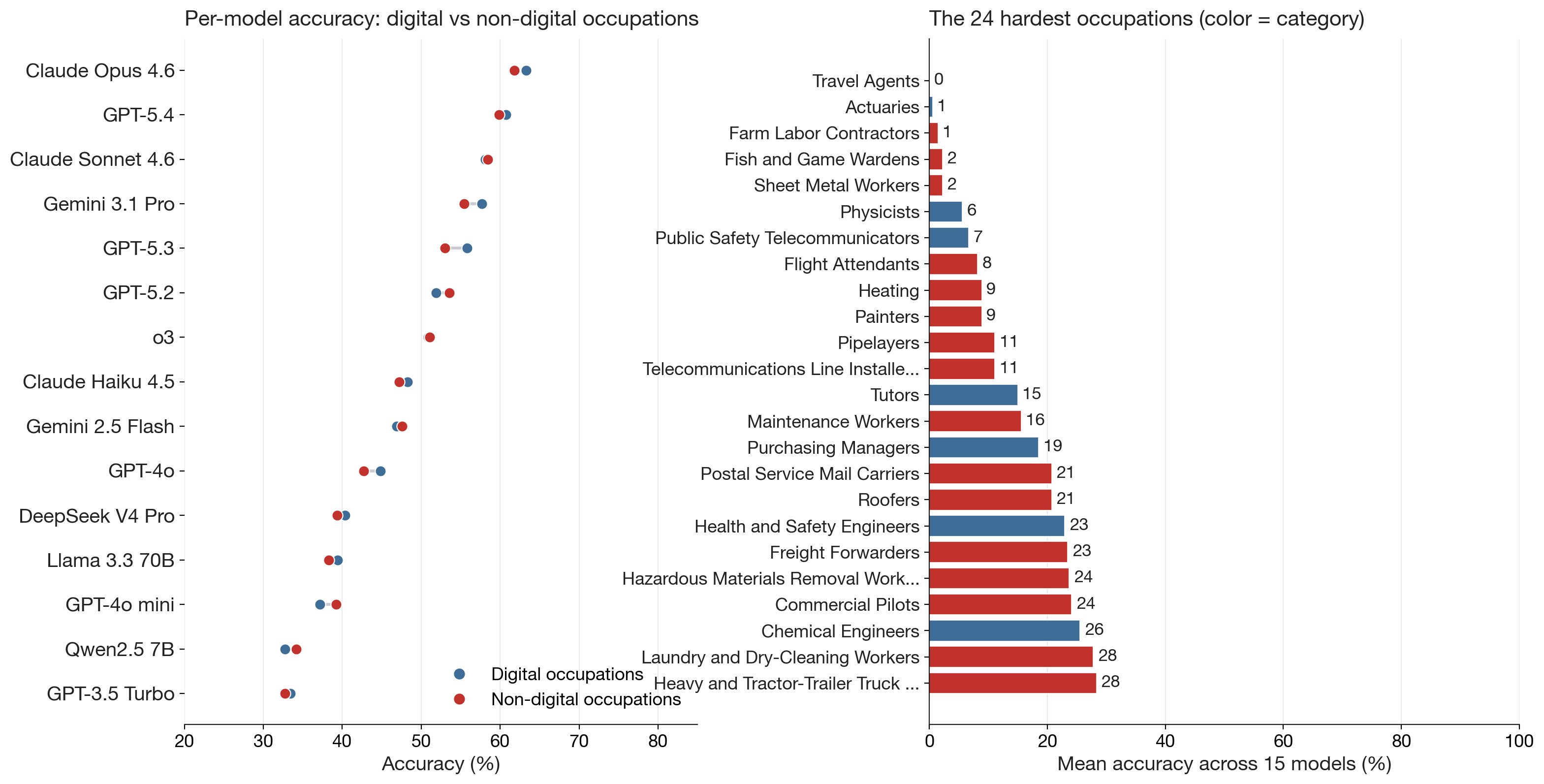}
  \figurenote{Occupations are hand-classified into digital (information and analytical desk work) and non-digital (physical, hands-on, in-person, clinical, or field work). Left: each model's mean accuracy on the two groups. Right: the twenty-four lowest-accuracy occupations, coloured by category.}
\end{figure}

\paragraph{Digital vs. non-digital jobs}
We also perform a final job level split based on whether the job is digital or non-digital. Digital jobs are defined as analytical or information-intensive office based jobs. Non-digital jobs are defined as field, in person, clinical, or hands on jobs. We find that performance is fairly even between digital and non-digital jobs (48.1\% of jobs in the digital category vs. 47.6\% in the non- digital category on average across models). Both digital and non-digital contain some of the most challenging jobs (Figure~\ref{fig:digital_nondigital}).

\subsection{Diagnosis of Pre-training Contamination in Closed-Book Tests}\label{appdx:contamination}

A potential issue in closed-book tests against publicly available authoritative sources is that frontier models may have encountered the same sources during pre-training and can select the correct answer via string-level rather than professional knowledge. We mitigate this by performing model- level splits on accuracy based on whether a source for a question appears after a given model's pre- training date. By definition, questions with sources after a given model's pretraining date are free from contamination for that model. If memorization were causing overestimation of ORQA's performance then the accuracy on post-cutoff sources should be consistently lower than pre-cutoff sources.

\paragraph{Mechanism for source-date assignment.}
We attempt to determine the publication date of a source from its URL using host-specific heuristics. We use the years in path elements like /2024/ and the fiscal year tag /fy23/ in URLs for OSHA grants. For content published on a low frequency but with URLs that do not contain date information (EPA, FAA, large professional society sites, OSHA directives, CDC stacks repository, BLS Occupational Outlook Handbook), we use a conservative prior date at the host level. URLs that do not conform to either heuristic are dropped. 86 (17.9\%) items are dated via URL pattern, 133 (27.7\%) items are dated via host-level prior and 261 (54.4\%) items are undatable and excluded. 219 dated items form the basis of the analysis.

\paragraph{Per-model cutoffs.}
The cutoff dates used for pre-training are derived from the most recent published model card for each provider or, if a cutoff is not publicly provided, from the conservative cutoff implied by the model release date: GPT-3.5 Turbo (Sep 2021), GPT-4o / GPT-4o-mini / o3 (Oct 2023), Llama 3.3 70B (Dec 2023), DeepSeek V4 Pro (Jul 2024), GPT-5.2 / GPT-5.3-chat-latest (Sep 2024), Qwen 2.5 7B (Oct 2024), Gemini 2.5 Flash (Dec 2024), Claude Opus 4.6 / Sonnet 4.6 / Haiku 4.5 (Jan 2025), Gemini 3.1 Pro (Mar 2025) and GPT-5.4 (Jun 2025).

\begin{figure}[!htbp]
  \centering
  \caption{Post-cutoff source items do not show the accuracy drop expected from pre-training memorization.}
  \label{fig:contamination}
  \includegraphics[width=\linewidth]{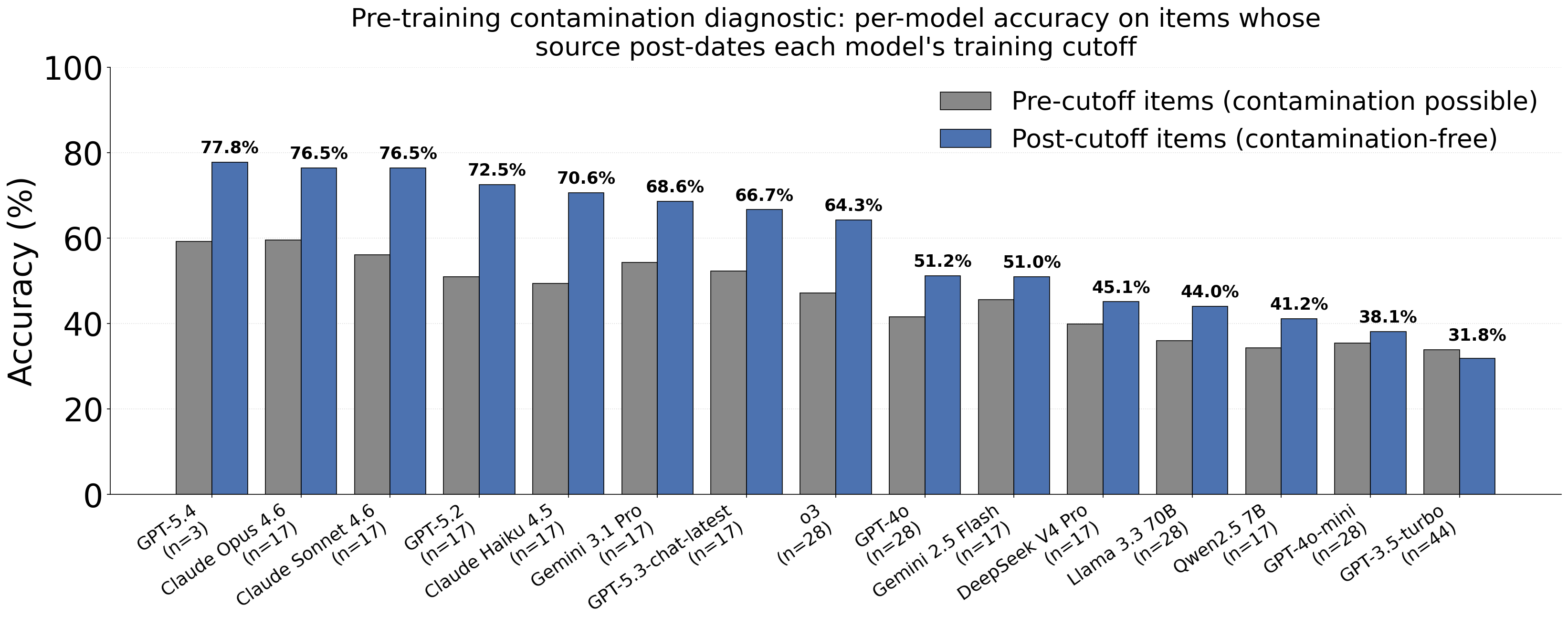}
  \figurenote{Source dates combine URL-derived dates and conservative host-level priors; undated items are excluded from this diagnostic. Post-cutoff samples are small for the newest models, so this is a directional contamination check.}
\end{figure}

\paragraph{Result.}
Figure~\ref{fig:contamination} shows per-model performance on pre- and post-cutoff strata. Generally performance is equal or higher on the post-cutoff strata than on the pre-cutoff strata: GPT-5.4 achieves 77.8\% performance on its (small) post-cutoff stratum (n = 3) compared to 59.3\% pre-cutoff (+18.5 pp), Claude Opus 4.6 achieves 76.5\% compared to 59.6\% (+16.9 pp), Claude Sonnet 4.6 achieves 76.5\% compared to 56.1\% (+20.4 pp), and GPT-5.2 achieves 72.6\% compared to 51.0\% (+21.6 pp). Accuracy due to contamination during pre-training seems unlikely given that there is not a systematic pre > post result across all fifteen models. Rather, fourteen of the fifteen models have post-cutoff accuracy equal to or greater than pre-cutoff accuracy (the exception is GPT-3.5 Turbo which has $-$2.1 pp). This suggests the diagnostic is more likely due to transfer of professional knowledge rather than exact knowledge from pre-training. We also note that the post-cutoff samples for the most recent frontier models are relatively small (n = 3 for GPT-5.4). This results in large statistical uncertainty when comparing pre- vs. post-cutoff performance for these models. A solution to this is to create an after-cutoff sub-bank as described below.

\paragraph{Future work: an after-cutoff sub-bank.}
The above diagnostic suffers from relatively small post-cutoff samples for the most recent frontier models. For a future iteration of the bank, we plan to create an after-cutoff sub-bank which only includes source materials that are after the cutoff time used for training of the most recently assessed model at the time of the sub-bank construction. This will allow for a contamination-free test without use of source-date heuristics and will increase the size of the post-cutoff sample which in turn should allow for tighter per-model confidence intervals. We already have support for source selection over a date range, so this is not a change in methodology but a targeted re-run.

\subsection{Prompts, Models, and Inference Settings}\label{appdx:prompts}

\paragraph{Evidence card and item creation prompts and quality check configuration.}
The prompts used for evidence card extraction, item creation, and all quality checks are provided in the supplementary material in their entirety. The prompt for evidence card extraction takes a JSON array of cards. Each card consists of a source excerpt (one to three sentences), a task or procedure from the excerpt, an occupation domain, and a quality rating. The item-builder prompt specifies that there must be six answer choices, that All of the above and None of the above must be present, that only one answer must be correct, and that the correct answer must be entailed by the source excerpt.

\paragraph{Details for each quality check.}
The quality checks described in Section~\ref{sec:method} are, in order: (1) source entailment, which verifies that the correct answer is entailed by the source excerpt. (2) question stem sanity, which verifies that the stem is professionally presented as a well-constructed question. (3) distractor plausibility, which verifies that all distractors are domain-plausible. (4) no distractor paraphrases another answer, which verifies that no distractor paraphrases another answer. (5) only one answer is correct, which verifies that exactly one of the six answers is correct. (6) Length and structural sanity: option lengths are within a factor of two of each other; no option is longer than 350 characters; and no phrases that indicate the source of knowledge. (7) Closed-book difficulty is estimated using Claude Haiku 4.5, Gemini 2.5 Flash and GPT-4o mini in parallel; any questions that can be answered by all three models are deemed too easy. (8) Eliminability is determined by an external LLM judging if each incorrect answer can be eliminated using only general world knowledge. (9) Auditing on concreteness checks for vagueness in the correct answer if any incorrect answer refers to a specific value or tool, and will then rewrite the correct answer to include the concrete information from the source. All of these are passed through to a categorization step that was tuned using human labels (Section~\ref{sec:method}).

\paragraph{Details on models and inference.}
We evaluated the following models: GPT-5.4, GPT-5.3-chat-latest, GPT-5.2, o3, GPT-4o, GPT-4o mini, GPT-3.5 Turbo, Claude Opus 4.6, Claude Sonnet 4.6, Claude Haiku 4.5, Gemini 3.1 Pro, Gemini 2.5 Flash, Llama 3.3 70B, DeepSeek V4 Pro and Qwen 2.5 7B. For each model we record the API version, temperature, and random seed used for the three inferences. Open-weight models are available via Together AI.

\paragraph{Models on the pipeline: generator and verifier}
For the bank construction pipeline there is a fixed set of generator and verifier models used across all items. On the generator side (OpenAI): item creation and evidence card extraction use gpt-4o (temperature = 0) and search/discovery and planning uses gpt-4o-mini. On the verifier side (separate from the generator, using Anthropic): the separate verifier, claude-haiku-4-5 (temperature = 0), performs the checks for: leakage, ambiguity, alignment, entailment, eliminability and concrete detail. The source authority check is implemented entirely in code using a whitelist per occupation and tier attribute. The only multi-model check is the difficulty pretest that runs in parallel gemini-2.5-flash (Google), claude-haiku-4-5 (Anthropic) and gpt-4o-mini (OpenAI). There are three models across three providers in order to access three separate rate limits and to achieve some independence in the errors. The use of separate providers for the generator (OpenAI), independent verifier (Anthropic), pretest (OpenAI, Anthropic and Google) justifies the use of the term independent verifier and also provides some resilience to self consistency error in the GPT-4o generator.

\subsection{External Validation Using External Signals}\label{appdx:validation}

\begin{figure}[!htbp]
  \centering
  \caption{External benchmark rankings and usage exposure provide directional validation for ORQA.}
  \label{fig:validation}
  \includegraphics[width=\linewidth]{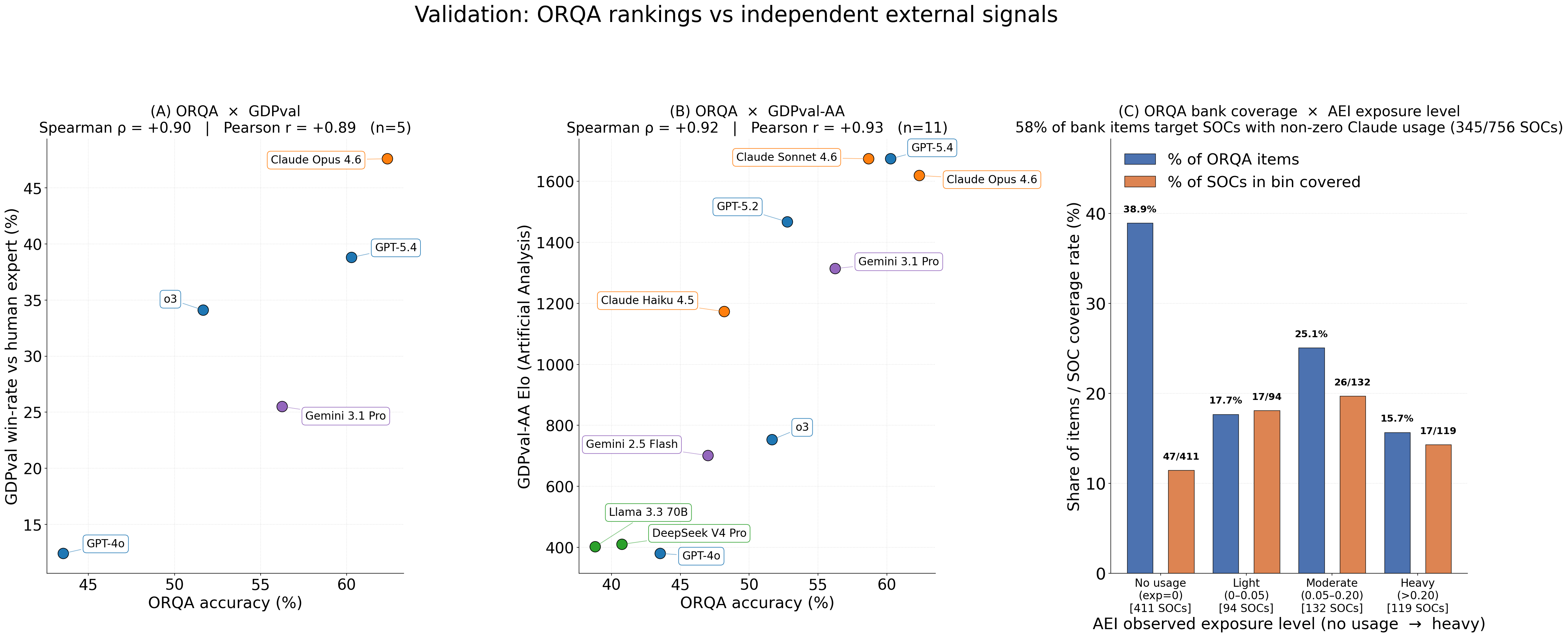}
  \figurenote{Panels compare ORQA with GDPval, GDPval-AA, and Anthropic Economic Index exposure where model or SOC coverage overlaps. These are triangulation checks for ranking and coverage, not claims that any external signal is the benchmark ground truth.}
\end{figure}

We provide a figure (Figure~\ref{fig:validation}) to complement the external validation presented in the main text. For the five models that overlap GDPval (\hyperref[fig:validation]{panel a}), there is a Spearman $\rho$ = +0.90 and a Pearson r = +0.89 correlation between the accuracy of ORQA and the performance of GDPval against industry- professionals. For the eleven models that overlap the GDPval-AA leaderboard for Artificial Analysis (\hyperref[fig:validation]{panel b}) there is a Spearman $\rho$ = +0.92 and a Pearson r = +0.93. \hyperref[fig:validation]{Panel (c)} compares the coverage of ORQA to the observed exposure of occupations to Claude usage in the Anthropic Economic Index. The 6-digit SOC codes are split into heavy, moderate, light, and no usage categories. 38.9\%, 17.7\%, 25.1\%, and 15.7\% of ORQA questions fall into the no usage, light, moderate, and heavy categories respectively. 61.1\% of bank questions use SOC codes that have non- zero observed exposure to Claude. ORQA covers 11\%, 18\%, 20\%, and 14\% of the SOC codes in each of these categories. This suggests the bank covers both frequently and infrequently used jobs beyond just those currently being adopted by AI. We also compare agreement per occupation to GDPval.

\paragraph{Per-occupation agreement with GDPval.}
GDPval also provides per-occupation performance information. This information is only provided in a small visualization within the GDPval site and not in a tabular form. Nine of the GDPval occupations are also in the current ORQA bank. We report GDPval win-rate in coarse buckets High / Mid / Low (High $\ge$50\%, Mid 25-49\%, Low < 25\%), and ORQA closed-book accuracy (averaged across fifteen models and three random seeds) for these. There is not exact agreement, but there is directional agreement. We highlight two points of disagreement. Nurse Practitioners are in the GDPval-Low bucket but achieve 82.8\% on ORQA. This may be due to the use of evidence from CDC and state board sources for the nurse practitioner questions in ORQA, which enables strong closed-book performance. Real Estate Brokers are in the GDPval-Mid bucket but only achieve 28.4\% on ORQA. Lawyers are in the GDPval-High bucket but only achieve 42.2\% on ORQA. This may be due to the differences in output expected by ORQA (identification of the specific provision entailed by the source) and GDPval (ability to produce long form responses). We note that the ORQA scores per occupation in Table~\ref{tab:gdpval_occ_overlap} are based on between 3 and 9 questions per cell. Thus, the scores within a cell are noisy. Table~\ref{tab:gdpval_occ_overlap} is intended as a qualitative sanity check, not an exact match.

\begin{table}[!htbp]
  \centering
  \caption{Per-occupation cross-check between ORQA and GDPval on the nine occupations present in both. \emph{Items} is the number of ORQA items contributing to the per-occupation accuracy. \emph{ORQA acc.}\ is the mean over fifteen models $\times$ three seeds. \emph{GDPval bucket} is a coarse three-way classification of GDPval win-rate against industry-professional experts (High $\geq 50$\%, Mid 25--49\%, Low $<25$\%). Rows are sorted by ORQA accuracy descending.}
  \label{tab:gdpval_occ_overlap}
  \small
  \begin{tabular}{lrrl}
    \toprule
    Occupation & Items & ORQA acc.\ & GDPval bucket \\
    \midrule
    Industrial Engineers          & 3 & 94.1\% & Mid  \\
    Nurse Practitioners           & 4 & 82.8\% & Low  \\
    Mechanical Engineers          & 8 & 67.5\% & Mid  \\
    Software Developers           & 3 & 65.9\% & Mid  \\
    Personal Financial Advisors   & 8 & 64.2\% & High \\
    Financial Managers            & 3 & 57.0\% & Mid  \\
    Lawyers                       & 6 & 42.2\% & High \\
    Real Estate Sales Agents      & 3 & 32.6\% & High \\
    Real Estate Brokers           & 9 & 28.4\% & Mid  \\
    \bottomrule
  \end{tabular}
\end{table}

\subsection{Interactive Dashboard}\label{appdx:dashboard}

A dashboard is provided online that enables users to view per-model and per-occupation results, filter by source domain and source category, and click through to individual items and their source texts. This enables the user to examine the evaluation without re-running any of the models while also providing access to the underlying data for further use.

\subsection{Full Per-Occupation Heatmap}\label{appdx:heatmap}

The per-occupation heatmap used in the paper (Figure~\ref{fig:occ_heatmap_quartile}) displays 45 occupations sampled from the 106 that each have at least three items, ordered by mean accuracy across the models and grouped into highest, median and lowest bands so that easy and hard domains are visually divided. In the released data, there are 116 occupations, some having only one or two items. The trends seen in the displayed heatmap also hold across the full set.

\subsection{Community-Thread Version}\label{appdx:community}
\label{sec:community}

We also created and tested a parallel community-thread version of ORQA. In this version, the correct answers are community-upvoted answers from public discussions about occupations instead of answers from authoritative sources. Authoritative answers describe what professionals should do, while community threads describe what professionals in fact ask when they encounter a problem at work. These two versions describe different dimensions of real-world professional consensus. This version is reported separately (and created separately) for completeness. It is not part of the 480-item authoritative dataset discussed above.

\paragraph{Pipeline.}
The community-thread pipeline begins with the list of occupations from O*NET. It groups jobs using SOC codes and retains one job per group. For each job, it searches public discussion forums for relevant community discussions (e.g., Stack Exchange, Metafilter, and other sites dedicated to occupations). Then, using GPT-4o, each community post is transformed into a multiple choice question with six possible answers. This version attempts to follow the style of queries that professionals might make in the real world (e.g., in WildBench \citep{lin2024wildbench}) but adapted to occupations. The community top-voted answer serves as the correct answer, other substantive answers are derived from the community post, and All of the above and None of the above are always included as answer choices. A four stage filtering system is used to improve on assistance framing, semantic relevance, plausibility of distractors, and the level of agreement in the community.

\paragraph{Final bank.}
The community-thread bank has 1,444 items in 758 occupations. Deduplication was performed based on SOC codes and these are nearly complete with respect to the O*NET list. Thus there are approximately 1.91 items per occupation. Having only two items in an occupation is allowed because this version of the system prioritizes coverage of occupations over within-occupation statistical power.

\begin{figure}[!htbp]
  \centering
  \caption{The community-thread variant preserves the frontier cluster under forum-derived ground truth.}
  \label{fig:community_leaderboard}
  \includegraphics[width=0.95\linewidth]{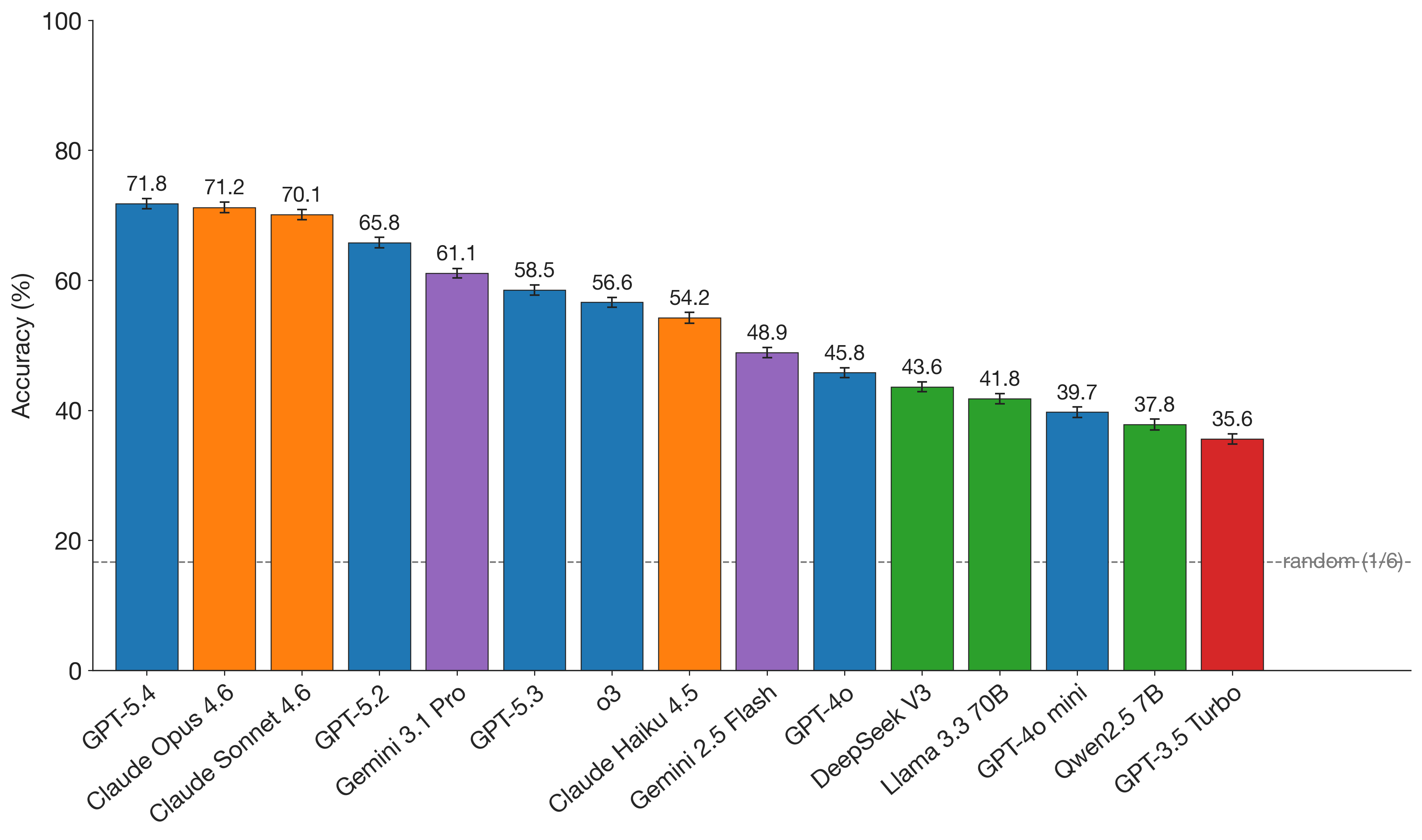}
  \figurenote{The community variant uses forum-derived ground truth over a separate 1{,}444-item, 758-occupation dataset. Scores are a complementary stress test and are not directly interchangeable with the authoritative-source ORQA bank.}
\end{figure}

\paragraph{Leaderboard.}
The closed-book leaderboard for the community-thread bank is shown in Figure~\ref{fig:community_leaderboard}. The frontier cluster is present in both the authoritative and community banks. Performance is more accurate on the community bank because community items are generally more answerable using general domain knowledge than items from authoritative regulations.

\begin{figure}[!htbp]
  \centering
  \caption{Community-thread performance varies substantially across occupations.}
  \label{fig:community_heatmap}
  \includegraphics[width=0.95\linewidth]{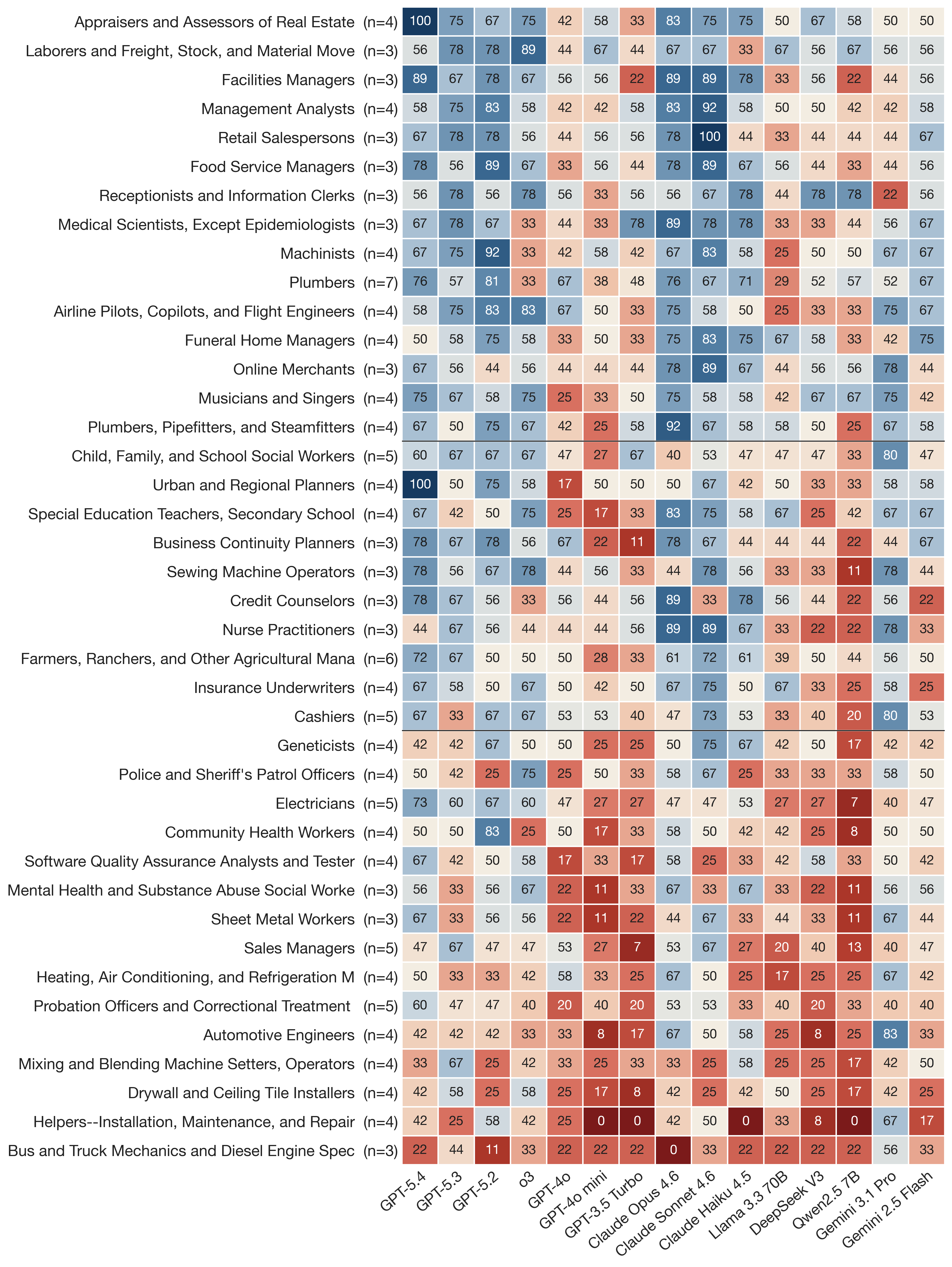}
  \figurenote{Displayed rows are sampled for readability from the top, middle, and bottom of the community-thread distribution. Occupation-level cells with few items should be read as diagnostics, not stable estimates.}
\end{figure}

\begin{figure}[p]
  \centering
  \caption{Community-thread occupation-by-model heatmap across the occupations with at least three items.}
  \label{fig:community_heatmap_all}
  \includegraphics[width=\linewidth, height=0.92\textheight, keepaspectratio]{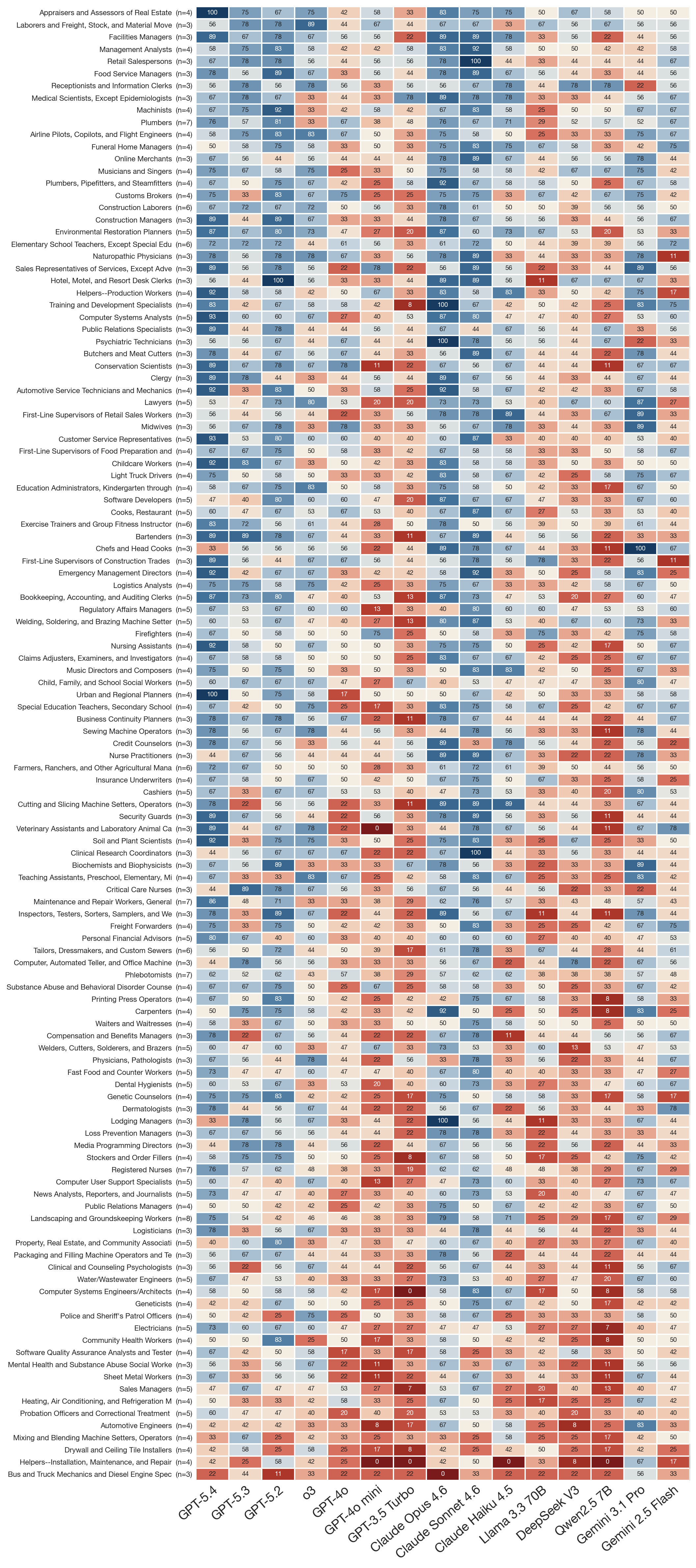}
  \figurenote{Rows are the community-thread occupations with at least three items, sorted by cross-model mean accuracy; columns are the fifteen evaluated models. Occupations with fewer than three items are omitted so that each cell rests on a more stable per-occupation estimate.}
\end{figure}

\paragraph{Per-occupation heatmap.}
A stratified per-occupation heatmap for the community-thread bank is shown in Figure~\ref{fig:community_heatmap}. Occupations were selected to sample across the full range of cross-model average correctness and are ordered from highest to lowest. A heatmap containing all occupations with three or more items is provided in Figure~\ref{fig:community_heatmap_all}.

\begin{figure}[!htbp]
  \centering
  \caption{SOC-level community-thread results summarize model performance by occupational family.}
  \label{fig:community_cluster}
  \begin{minipage}[c]{0.49\linewidth}
    \centering
    \includegraphics[width=\linewidth]{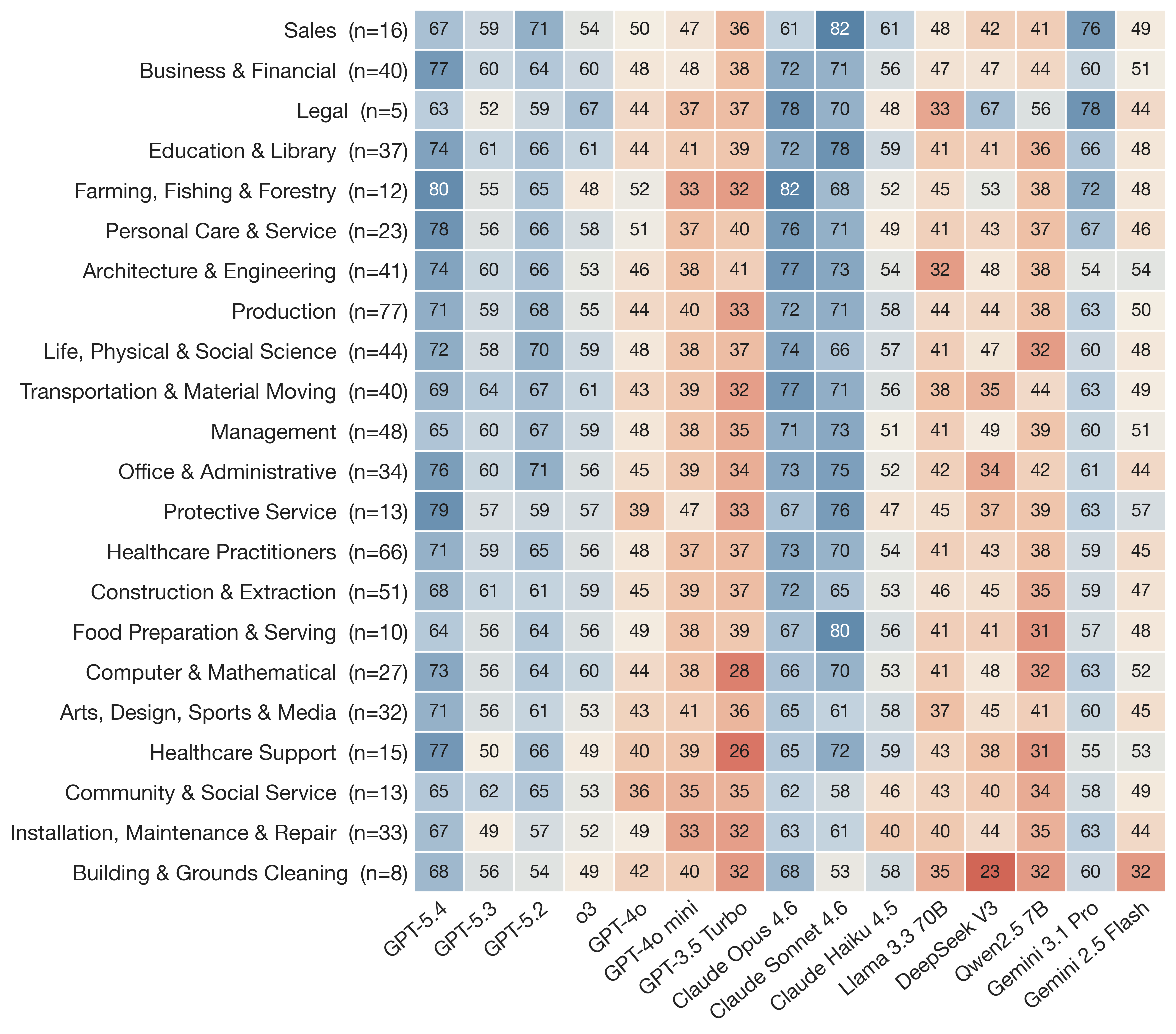}
  \end{minipage}\hfill
  \begin{minipage}[c]{0.49\linewidth}
    \centering
    \includegraphics[width=\linewidth]{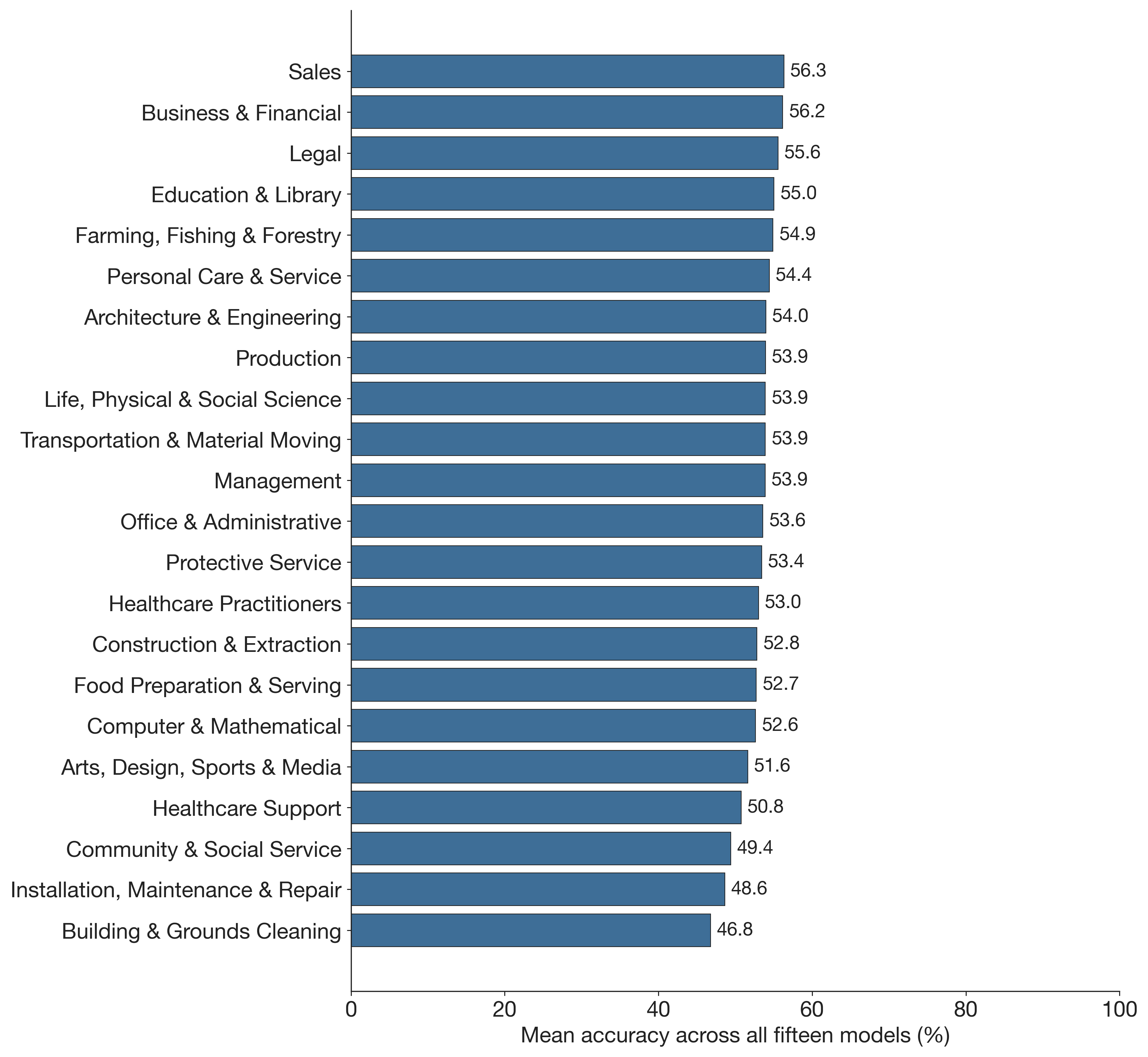}
  \end{minipage}
  \figurenote{The left panel shows SOC major-group accuracy by model; the right panel averages within each group and then across groups.}
\end{figure}

\paragraph{SOC major-group aggregation.}
Community-thread results are aggregated to SOC major groups in Figure~\ref{fig:community_cluster}. \hyperref[fig:community_cluster]{The left panel} is a cluster-level heatmap. \hyperref[fig:community_cluster]{The right panel} is a cluster-level leaderboard that computes group averages first within the major groups of SOC occupations and then across the major groups. All major occupation groups are given equal weight in the leaderboard. The ordering of the fifteen models remains the same across the leaderboard.

\begin{figure}[!htbp]
  \centering
  \caption{Community-thread leaderboards remain stable under labor-market weighting.}
  \label{fig:community_weighted}
  \includegraphics[width=0.95\linewidth]{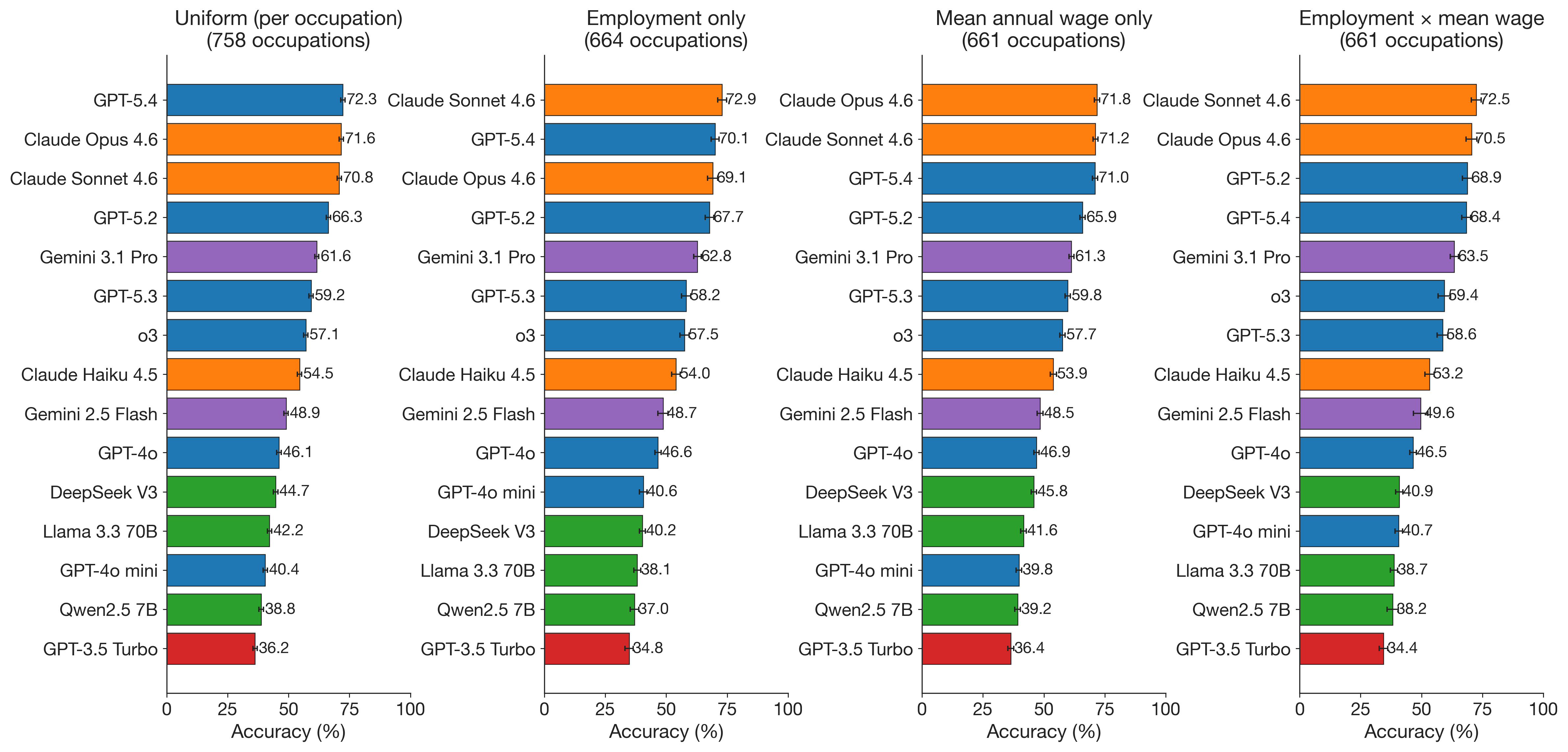}
  \figurenote{The three weighted variants assign occupation weights proportional to total employment, annual mean wage, and their product. The underlying model responses are unchanged; only each occupation's contribution to the final average varies.}
\end{figure}

\paragraph{Economic-importance-weighted leaderboards.}
Three economic-importance-weighted leaderboards are reported in Figure~\ref{fig:community_weighted} across the community- thread bank. Additionally, the unweighted leaderboard is also reported. The occupations are weighted according to the number of people employed in the nation, according to the mean wage per year, and according to the product of the first two. In all three cases, the ordering of the fifteen models remains the same.

\begin{figure}[!htbp]
  \centering
  \caption{Most community-thread runs show no strong answer-position bias.}
  \label{fig:community_answerdist}
  \includegraphics[width=0.95\linewidth]{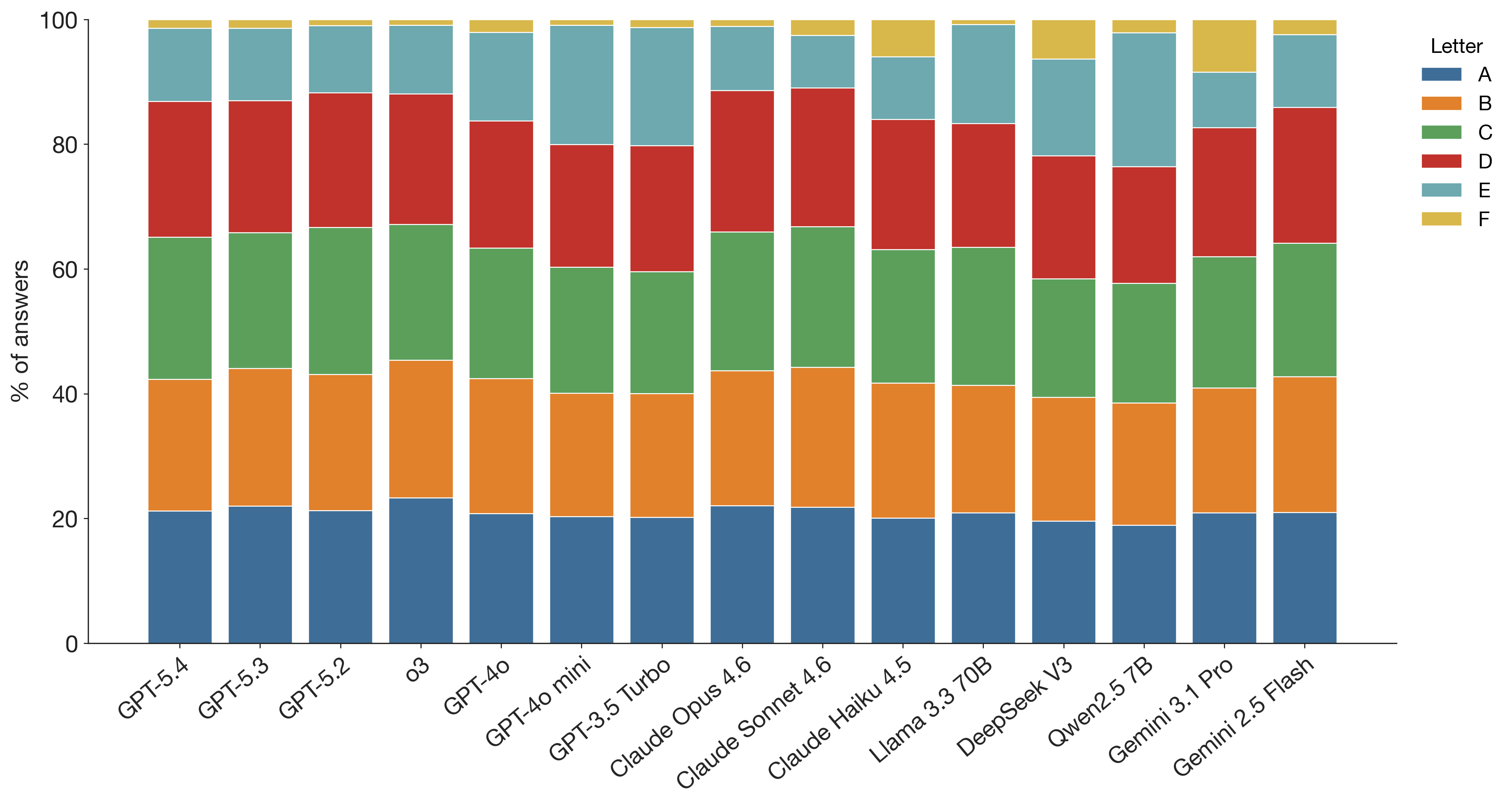}
  \figurenote{Options A--D are randomized substantive choices; E and F are the fixed \emph{All of the above} and \emph{None of the above} slots.}
\end{figure}

\paragraph{Positional bias and distribution of selected answers.}
Figure~\ref{fig:community_answerdist} reports the distribution of selected answers (A to F) in the community-thread bank. For the majority of models, answers are distributed fairly evenly across A-D (~20-23\% for each) and answer E (All of the above) is chosen about 10-13\% of the time. Answer F (None of the above) is chosen infrequently.

\paragraph{Full community per-occupation heatmap.}
Figure~\ref{fig:community_heatmap_all} shows the per-occupation heatmap of correctness for the community-thread answers for occupations that have at least three questions. The reliability criterion used in the authoritative per-occupation heatmap is also used here. Occupations with fewer than three questions are excluded from this heatmap because the per-occupation cells would be based on only one or two questions.

\FloatBarrier

\begin{table}[h]
\caption{Schema for the occupation roster table included in the supplementary material. The full table lists every occupation in the ORQA bank together with its SOC code and item count.}
\label{tab:occupations}
\centering
\begin{tabular}{llr}
\toprule
Occupation title & SOC code & Items \\
\midrule
Firefighters & 33-2011 & 11 \\
Hydrologists & 19-2043 & 10 \\
Real Estate Brokers & 41-9021 & 9 \\
Pest Control Workers & 37-2021 & 9 \\
Freight Forwarders & 43-5011 & 9 \\
\ldots & \ldots & \ldots \\
\bottomrule
\end{tabular}
\end{table}

\begin{table}[h]
\caption{Schema for the source roster table included in the supplementary material. The full table lists every authoritative source URL used to construct an item, together with source domain and category.}
\label{tab:sources}
\centering
\begin{tabular}{llll}
\toprule
Item ID & Source domain & Source URL & Category \\
\midrule
Q001 & osha.gov & osha.gov/laws-regs/\ldots & Federal \\
Q002 & amftrb.org & amftrb.org/\ldots & Professional / board \\
Q003 & nfpa.org & docinfofiles.nfpa.org/\ldots & Professional \\
Q004 & faa.gov & faa.gov/\ldots & Federal \\
\ldots & \ldots & \ldots & \ldots \\
\bottomrule
\end{tabular}
\end{table}

\end{document}